\documentclass[10pt,twocolumn,letterpaper]{article}

\usepackage{cvpr}              

\usepackage{graphicx}
\usepackage{amsmath}
\usepackage{amssymb}
\usepackage{booktabs}

\usepackage{times}
\usepackage{epsfig}
\usepackage[moderate]{savetrees}
\usepackage[table,xcdraw]{xcolor}
\usepackage{import}
\usepackage{mathrsfs}
\usepackage{adjustbox}
\usepackage{multirow}

\newcommand\numberthis{\addtocounter{equation}{1}\tag{\theequation}}

\usepackage[pagebackref,breaklinks,colorlinks]{hyperref}

\usepackage[capitalize]{cleveref}
\crefname{section}{Sec.}{Secs.}
\Crefname{section}{Section}{Sections}
\Crefname{table}{Table}{Tables}
\crefname{table}{Tab.}{Tabs.}

\def\cvprPaperID{10122} 
\def\confName{CVPR}
\def\confYear{2022}

\begin{document}

\title{Dissecting the impact of different loss functions with gradient surgery}

\author{Hong Xuan\\
Microsoft\\
{\tt\small Hong.Xuan@microsoft.com }
\and
Robert Pless\\
Geroge Washington University\\
{\tt\small pless@gwu.edu}
}
\maketitle

\begin{abstract}
   Pair-wise loss is an approach to metric learning that learns a semantic embedding by optimizing a loss function that encourages images from the same semantic class to be mapped closer than images from different classes.  The literature reports a large and growing set of variations of the pair-wise loss strategies. Here we decompose the gradient of these loss functions into components that relate to how they push the relative feature positions of the anchor-positive and anchor-negative pairs.  This decomposition allows the unification of a large collection of current pair-wise loss functions.  Additionally, explicitly constructing pair-wise gradient updates to separate out these effects gives insights into which have the biggest impact, and leads to a simple algorithm that beats the state of the art for image retrieval on the CAR, CUB and Stanford Online products datasets.
\end{abstract}

\section{Introduction}
Deep Metric Learning trains networks to map semantically related images to similar locations in an embedding space.  Metric learning is useful in extreme classification settings when there are so many classes and limited embedding size that standard approaches fail, when there is a need to compare features extracted from images in unseen classes, or when there may be incomplete labels that allow a system to know that images come from the same or different classes without knowing what those classes are.

In this domain, one popular pair-wise loss to train a network is Triplet Loss.  Triplets are three images comprising an anchor image, a positive image from the same class, and a negative image from a different class.  The network is trained with a loss function that penalizes situations where the anchor-negative pair is closer than the anchor-positive pair.  Many variations of this basic approach explore ways to choose which triplets should be included in the optimization or how much they should be weighted, whether the optimization should consider distances or angles between the embedded vectors, and what specific loss function should drive the scoring of a particular triplet.

One recent work gave a large-scale analysis of many of these variations~\cite{musgrave2020metric} and found that a substantial fraction of the reported performance variation disappears with careful matching of experimental conditions.  In this work, we propose a further unifying analysis of these approaches, but explicitly consider how the pair-wise loss function attempts to affect the embedding location of the anchor, positive, and negative examples.  Different pair-wise loss functions have gradients that directly affect the desired locations of each embedded location in different ways. Those gradients are different in terms of the {\it direction} the anchor, positive and negative examples are pushed, the {\it overall importance} or weight given to different triplets, and the {\it relative importance} or weight given to the anchor-positive vs. the anchor negative pairs.

\begin{figure}[t]
    \centering
    \includegraphics[width=1\columnwidth]{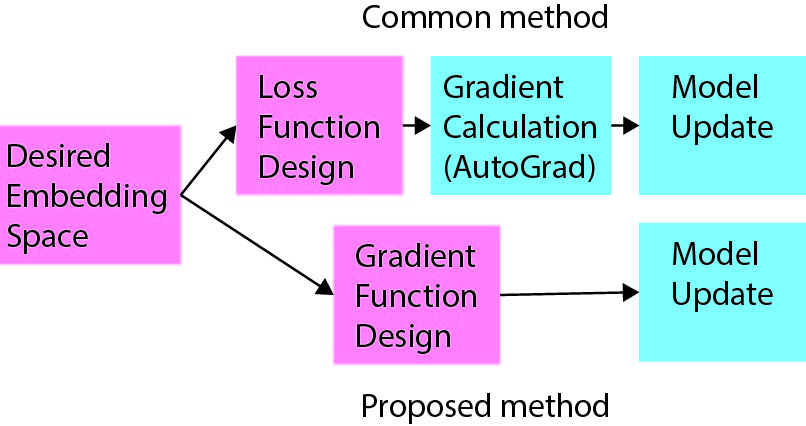}
    \caption{To realize a desired embedding space, a common method is to design a loss function which can be calculated on deep learning platforms such as PyTorch and TensorFlow(Red). The auto-grad mechanism on the platforms automatically calculates the gradient to update the model parameters to forming the desired embedding space(Blue). In practice, the goal of deep metric learning is about optimizing the separation or clustering of feature points extracted from imagery, and the loss function is a somewhat indirect approach to reach that goal, while the gradient more directly affects the update of the feature extraction.   We propose the method to directly design the gradient to train models.
    }
    \label{fig:main}
\end{figure}

In addition to the analysis, we exploit the fact that PyTorch~\cite{pytorch} allows for the programmatic specification of gradients, allowing us to explicitly control the above gradient components, and then supports back-propagation to encourage the low-level features to move in this way. This flexibility allows us to explore the relative contributions of these components of the gradient and better understand what is and is not important in the optimization. Finally, we demonstrate the potential to directly modify the gradient components to train models for deep metric learning tasks instead of loss function modification.

The three main contributions\footnote{Reject in CVPR2021, ICCV2021, CVPR2022} of this are:
\begin{itemize}
    \item a direct gradient framework to create a unified analysis of many recent triplet and pair-wise loss functions in terms of their gradients,
    \item an experimental analysis showing how different choices for components of the gradient affects model performance,
    \item a deeper understanding of the practical effects of defining a loss based on the Euclidean metric compared with the cosine similarity metric, and
    \item an integration of the best choice of each component to create a new gradient rule that outperforms the current state-of-art result for image retrieval by a clear margin across multiple datasets.
\end{itemize}

\section{Background}
There are many loss functions that have been proposed to solve the deep metric learning problem. Pair-wise loss functions such as contrastive loss~\cite{hadsell2006dimensionality}, binomial deviance loss~\cite{yi2014deep}, lifted structure loss~\cite{SOP} and multi-similarity loss~\cite{wang2019multi} penalize pairs of same label instances if their distance is large and pairs of different label instances if their distance is small.  The triplet loss function~\cite{hoffer2015deep,facenet} and its variants such as circle loss~\cite{Sun_2020_CVPR} form a triplet that contains anchor, positive and negative instances, where the anchor and positive instance share the same label, and anchor and negative instance share different labels. These loss functions have losses encouraging the anchor-positive distance to be smaller than anchor-negative distance.  Other variants of triplet loss integrate more negative instances into a triplet, such as N-Pair loss~\cite{Npairs}.  Proxy loss~\cite{Proxy} defines for each class a learnable anchor as a proxy. During the training, each instance is directly pulled to its proxy and pushed away from the proxy location for other classes.

Due to the explosion of many new loss functions, issues underlying the fair comparison for these loss functions have been raised in~\cite{musgrave2020metric}. This paper works hard to re-implement many works before 2019. It tries to fix settings such as network architecture, optimizer and image prepossessing and compares different methods apple to apple.  This gives a relatively clear comparison of many loss functions but does not try to explore why some methods are superior to others. 

Recent works such as Multi-Similarity Loss and Circle Loss~\cite{wang2019multi,Sun_2020_CVPR,Xuan_2020_ECCV} have started with standard triplet loss formulations and adjust the gradient of loss functions to give clear improvements with very simple code modifications.  These works all find an explicit loss function whose gradient creates the desired loss function.  In some cases, like the current state-of-the-art approach across many datasets~\cite{wang2019multi}, the updated loss function for one triplet includes the relative similarity between the anchor-positive and the anchor and other examples for the anchor's class.  This more complicated loss function and more complicated gradient may cause subtle challenges in the optimization process.

Other strategies start with a desired gradient weighting function and integrate the desired gradients to solve for a loss function whose gradient has the appropriate properties.  This is often limited to simple weighting strategies, such as the simple linear form in~\cite{Sun_2020_CVPR} and simple gradient removal for positive pairs when triplets contain hard negative in~\cite{Xuan_2020_ECCV}, because it may be hard to find the loss function whose gradient is consistent with complex weighting strategies.

The discussion of explicitly updating the direction of the gradient has been introduced in~\cite{Mohan_2020_CVPR}.  They encourage the anchor-positive and anchor-negative directional updates to be orthogonal (so they don't cancel each other), but include this as a  "direction regularization", which does not enforce orthogonality.

The most related work is P2Sgrad~\cite{zhang2019p2sgrad}, the author analyzes the gradient in the family of margin-based softmax loss and directly modified the gradient with the cosine similarity for better optimization. Comparing to P2Sgrad, our work focuses on the triplet and pair-wise loss functions.

The framework in this paper directly explore the space of desired gradient updates as shown in Figure~\ref{fig:main}.  By not limiting ourselves to designing a loss function with appropriate gradients, we can be more explicit in experimentally dissecting the effects of different parts of the gradient.  Furthermore, we can recombine the gradient terms that are experimentally most useful in a form of gradient surgery~\cite{yu2020gradient} that very slightly alters existing algorithms to give improved performance.

\section{The Role of the Gradient in Metric Learning}
We define a collection of terms for how a batch of images affects a network.
Let $\mathbf{X}$ be a batch of input images, $\mathbf{f}$ be the $L2$ normalized feature vectors of the images extracted by the network, 
$l$ be loss value for the batch, 
$\theta$ be the parameters of the network, 
$\eta$ be the learning rate,
 $f_{\theta}(\cdot)$ be the mapping function of the network, and $L(\cdot)$ be loss function. In the forward training step, the expression is:
 
\begin{equation}
l=L(\mathbf{f}) \text{,  where  } \mathbf{f}=f_{\theta}(\mathbf{X})
\end{equation}. 

The network weights are updated as:
\begin{equation}
\theta^{t+1} = \theta^{t} - \eta\frac{\partial L}{\partial \mathbf{f}}\frac{\partial \mathbf{f}}{\partial \theta}
\end{equation}

This equation highlights that the gradient of the loss function (rather than the loss function itself) directly affects how the model updates its parameters. Therefore, explicitly exploring the gradient is a useful path to exploring network learning behavior.

We decompose the gradient into two terms, $\frac{\partial L}{\partial \mathbf{f}}$ and $\frac{\partial \mathbf{f}}{\partial \theta}$. The first term represents how changing the embedded feature location affects the loss, and this is the term explored most in detail in this work. The second term represents how model parameter (network weight) changes affect the feature embedding.  In a modern deep network with multiple layers, the second term is always expanded with the multiplication of multiple terms for each layer because of the derivative chain rule.

In the following discussion, we focus on the particular forms of the first term in many triplet and pair-wise loss functions and then proposed to directly set and design the first term for model training. In Section~\ref{sec:idea}, as an example, two commonly used triplet losses are decomposed into components and then those components are categorized into three parts. Then, Section~\ref{sec:idea_gd},~\ref{sec:idea_wp} and ~\ref{sec:idea_wt} extend the analysis to more existing loss functions.

\subsection{Gradient of Triplet Losses}
\label{sec:idea}
Given a triplet, $(\mathbf{f_a},\mathbf{f_p},\mathbf{f_n})$, there are two commonly used triplet losses in the literature, a triplet loss based on Euclidean distance:
\begin{equation}
L_{euc} = \max (D_{ap}^2-D_{an}^2+m, 0),
\label{eq:euc_loss}
\end{equation}
where $D_{ap}=\|\mathbf{f_a}-\mathbf{f_p}\|$, $D_{an}=\|\mathbf{f_a}-\mathbf{f_n}\|$ are the distances between the anchor-positive and the anchor-negative pairs, and $m$ is a distance margin. A second common triplet loss is the triplet loss based on cosine similarity with NCA~\cite{nca}:
\begin{equation} 
L_{cos}
=-\log(\frac{\exp{(\tau S_{ap})}}{\exp{(\tau S_{ap})}+\exp{(\tau S_{an})}})
\label{eq:nca_loss}
\end{equation}
where $S_{ap}=\mathbf{f_a}^{T}\mathbf{f_p}$, is the cosine similarity computed as the dot-product of the normalized anchor feature and the normalized feature from the positive example, the anchor-negative is computed in the same way, $S_{an}=\mathbf{f_a}^{T}\mathbf{f_n}$ and $\tau$ is the scaling parameter.

When comparing these two loss functions, their substantial differences make it challenging to determine how the loss affects performance. One loss is based on the Euclidean distance combined with a hinge function, while the other uses cosine similarity along with a negative log softmax function to combine the anchor-positive and anchor-negative pairs.  Looking at the gradients of these loss functions makes the difference more clear.  In triplet loss based on Euclidean distance, if the loss is greater than 0, its gradient can be derived from Equation~\ref{eq:euc_loss} as:
\begin{equation}
\left\{
\begin{aligned}
&\frac{\partial L_{euc}}{\partial \mathbf{\mathbf{f_p}}}
= 2\|\mathbf{f_p}-\mathbf{f_a}\|\mathbf{e_{p}^{euc}}\\
&\frac{\partial L_{euc}}{\partial \mathbf{\mathbf{f_n}}}
= 2\|\mathbf{f_a}-\mathbf{f_n}\|\mathbf{e_{n}^{euc}}\\
&\frac{\partial L_{euc}}{\partial \mathbf{\mathbf{f_a}}}
=-2\|\mathbf{f_a}-\mathbf{f_p}\|\mathbf{e_{p}^{euc}}-2\|\mathbf{f_n}-\mathbf{f_a}\|\mathbf{e_{n}^{euc}}
\end{aligned} 
\right.
\label{eq:gradient_margin}
\end{equation}
Being explicit about this gradient allows us to name the direction that the positive example is being pulled to anchor example as $\mathbf{e_{p}^{euc}}$, and these are unit vectors defined as: $\mathbf{e_{p}^{euc}}=\frac{\mathbf{f_p}-\mathbf{f_a}}{\|\mathbf{f_p}-\mathbf{f_a}\|}$, with corresponding directions for the negative example, $\mathbf{e_{n}^{euc}}=\frac{\mathbf{f_a}-\mathbf{f_n}}{\|\mathbf{f_a}-\mathbf{f_n}\|}$. 

The gradient of the triplet loss function based on cosine similarity can also be derived from Equation~\ref{eq:nca_loss} to give a unit direction and magnitude:
\begin{equation}
\left\{
\begin{aligned}
&\frac{\partial L_{cos}}{\partial \mathbf{f_p}} 
=\frac{1}{1+\exp{(\tau(S_{ap}-S_{an}))}} \tau \mathbf{e_p^{cos}}\\
&\frac{\partial L_{cos}}{\partial \mathbf{f_n}} 
=\frac{1}{1+\exp{(\tau(S_{ap}-S_{an}))}} \tau \mathbf{e_n^{cos}}\\
&\frac{\partial L_{cos}}{\partial \mathbf{f_a}} 
=\frac{1}{1+\exp{(\tau(S_{ap}-S_{an}))}} \tau (\mathbf{e_{a_p}^{cos}}+\mathbf{e_{a_n}^{cos}})
\end{aligned} 
\right.
\label{eq:gradient_nca}
\end{equation}
where $\mathbf{e_p^{cos}}=-\mathbf{f_a}$, $\mathbf{e_n^{cos}}=\mathbf{f_a}$, $\mathbf{e_{a_p}^{cos}}=-\mathbf{f_p}$ and  $\mathbf{e_{a_n}^{cos}}=\mathbf{f_n}$ are the unit gradient directions.

Though both $L_{euc}$ and $L_{cos}$ contain different gradient components, those components can be categorized into two major parts: unit gradient direction for moving the feature and a scalar weight that affects the length of the gradient in that direction.  The weight itself can be divided into two sub-parts: the weight related to all three features in a triplet $\mathbf{f_a}$, $\mathbf{f_p}$ and $\mathbf{f_n}$ (\textbf{Triplet Weight}), and the weight related to the positive pair $\mathbf{f_a}$ and $\mathbf{f_p}$ or negative pair $\mathbf{f_a}$ and $\mathbf{f_n}$ in a triplet (\textbf{Pair Weight}).

With the categorizations of the gradient components, it becomes easy to compare the effects of each component. Before the comparison, we first show how recently proposed loss functions can be characterized by computing the direction and weights of the different gradient terms 
in Sections~\ref{sec:idea_gd},~\ref{sec:idea_wp} and~\ref{sec:idea_wt} and then perform comparisons of the isolated effects of each gradient component in Sections~\ref{sec:exp_gd},~\ref{sec:exp_wp} and~\ref{sec:exp_wt}

\subsection{Unit Gradient Direction}
\label{sec:idea_gd}
\begin{figure}[t]
    \centering
    \includegraphics[width=1.0\columnwidth]{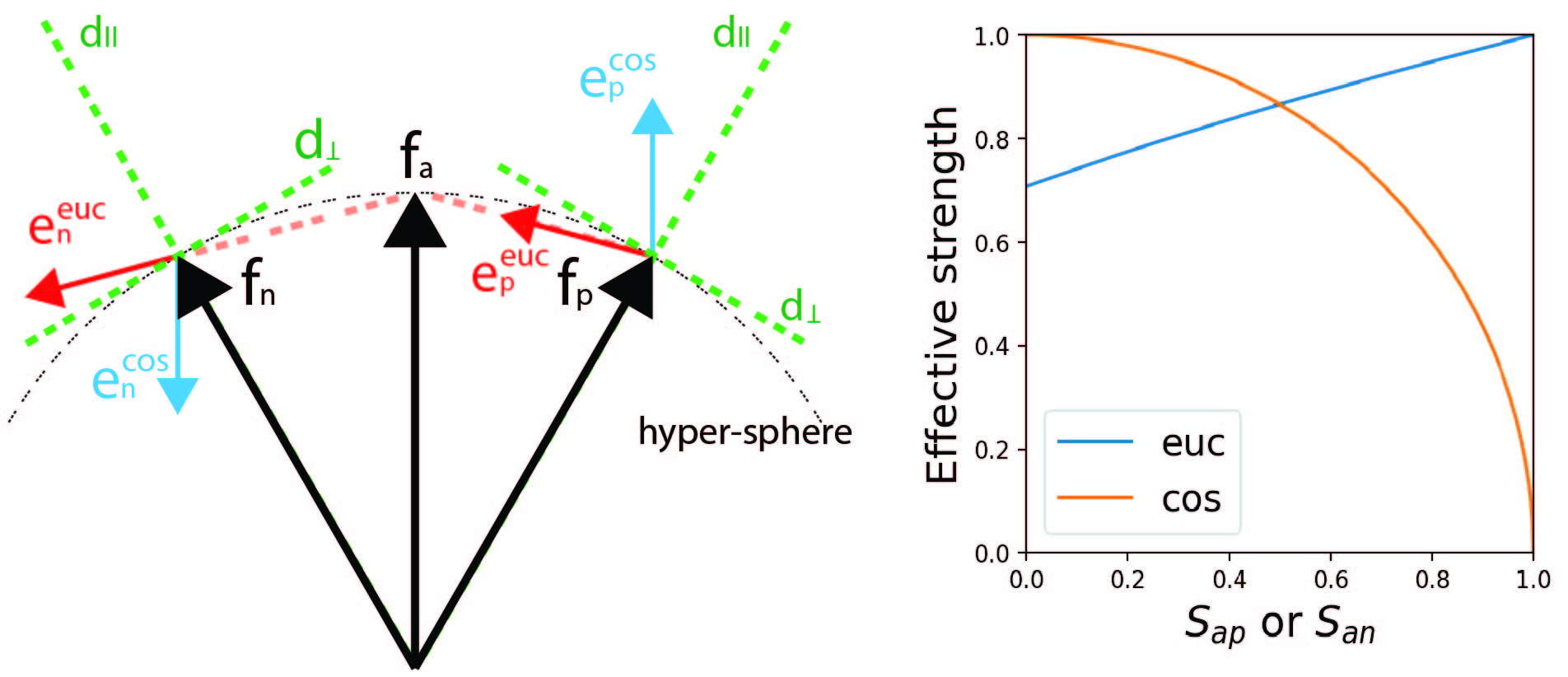}
    \caption{Left: Illustration of two types of gradient direction $\mathbf{e^{euc}}$ (red) and $\mathbf{e^{cos}}$ (blue), and two gradient decomposition directions (green). Right: Gradient projection strength along $d_{\bot}$ of $\mathbf{e^{euc}}$ and $\mathbf{e^{cos}}$ varies with the similarity of positive/negative pairs.}
    \label{fig:cos_euc_thoery}
\end{figure}

The first gradient component is the unit vector in the direction of the gradient, derived from how the loss function moves the relative configuration of the anchor, positive and negative features.  We refer to the unit gradient direction of the two most common metrics Euclidean distance and cosine similarity as Euclidean direction $\mathbf{e^{euc}}$ and cosine direction $\mathbf{e^{cos}}$.   Recent work~\cite{Mohan_2020_CVPR}, also suggests to other directions, Euclidean orthogonal direction $\mathbf{e^{euc-orth}}$ and cosine orthogonal direction $\mathbf{e^{cos-orth}}$.

\paragraph{Euclidean Direction($\mathbf{e^{euc}}$):}
In equation~\ref{eq:gradient_margin}, the geometric explanation of Euclidean direction is to move the positive feature directly towards the anchor and move the negative feature directly away from the anchor, as shown in Figure~\ref{fig:cos_euc_thoery}.  The vector direction of the anchor image (not shown in the Figure), is a combination of these directions.

\paragraph{Cosine Direction($\mathbf{e^{cos}}$):}
In equation~\ref{eq:gradient_nca}, the geometric explanation of cosine direction on positive pair is to move the positive feature in the anchor feature direction and move the anchor in positive feature direction, and on negative pair is to move negative in the opposite of the anchor feature direction and move the anchor in the opposite of the negative feature direction as shown in Figure~\ref{fig:cos_euc_thoery}.

\paragraph{Orthogonal Direction($\mathbf{e^{euc-orth}}$ $\&$ $\mathbf{e^{cos-orth}}$):} 
A direct gradient modification function~\ref{eq:gradient_orth} can be applied to both the  Euclidean and cosine directions.  This requires the negative pair to move in a direction orthogonal to the direction the positive pair is moving.  This is constrained as:
\begin{equation}
\mathbf{e_n}\cdot(\mathbf{f_a}-\mathbf{f_p})=0
\label{eq:gradient_orth}
\end{equation}

This gradient was realized in recent work by~\cite{Mohan_2020_CVPR} who implicitly encourage the negative examples to move orthogonally to the anchor positive direction by adding regularizer in their loss function.  Our approach is directly understanding the gradient direction for each example highlights the impact of this loss function.

\subsection{Pair Weight}
\label{sec:idea_wp}
We define the pair-weight $P$, for the anchor-positive pair $P_{+}$ and anchor-negative pair $P_{-}$. The pair weight of cosine similarity $P^{cos}$ based triplet loss is a constant scaling parameter.  This is useful as a baseline for comparison. For this case where both pair weights are set with constant $1$, as:
\begin{equation}
P_{+}^{con}=P_{-}^{con}=1;
\end{equation}

In Euclidean distance based triplet loss, the pair weight $P^{euc}$ is different for the anchor-positive and anchor-negative pairs:
\begin{equation}
\left\{
\begin{aligned}
&P_{+}^{euc}=\|\mathbf{f_a}-\mathbf{f_p}\|\\
&P_{-}^{euc}=\|\mathbf{f_a}-\mathbf{f_n}\|
\end{aligned} 
\right.
\end{equation}
and indicates the pair weight is proportional to the distance between the anchor and the other element of the pair.

Recent works~\cite{Sun_2020_CVPR, wang2019multi, Xuan_2020_ECCV} argue that the weight for anchor-negative pair should be large when they are close to each other. Otherwise, as mentioned in ~\cite{Xuan_2020_ECCV}, the optimization will quickly converge to bad local minima. The solution in Circle loss~\cite{Sun_2020_CVPR} is to apply a linear pair weight $P^{lin}$: for negative pairs, the weight is large if the similarity is large and small if the similarity is small; for positive pairs, the weight is large if the similarity is small and small if the similarity is large:
\begin{equation}
\left\{
\begin{aligned}
&P_{+}^{lin}=1-S_{ap}\\
&P_{-}^{lin}=S_{an}
\end{aligned} 
\right.
\label{eq:wp_lin}
\end{equation}

Early work binomial deviance loss~\cite{yi2014deep} uses a similar pair weight but with a nonlinear sigmoid form $P^{sig}$:
\begin{equation}
\left\{
\begin{aligned}
&P_{+}^{sig}=\frac{1}{1+\exp{( \alpha(S_{ap}-\lambda))}}\\
&P_{-}^{sig}=\frac{1}{1+\exp{(-\beta (S_{an}-\lambda))}}
\end{aligned} 
\right.
\label{eq:wp_sig}
\end{equation}
where $\alpha$, $\beta$ and $\lambda$ are three hyper-parameters. 

Multi-similar(MS) loss~\cite{wang2019multi} combines ideas from the lifted structure loss~\cite{SOP} and binomial deviance loss~\cite{yi2014deep}, which includes not only the self-similarity of a selected pair but also the relative similarity from other pairs.  

The MS paper~\cite{wang2019multi} tries to find a loss function whose derivative fits the proposed pair weight.  Because the relative similarity term involves additional examples (outside the triplet), this creates additional gradients relative to those examples, even though the stated purpose is to weigh the selected pair.  Therefore, it's difficult to understand if the performance gain is coming from the proposed pair weight or from the gradients affecting the feature location of these other examples.  By casting their work within our framework, we can decouple the pair-weighting and explore the impact of this term in isolation.

We follow the MS paper to cast their weighting function $P^{sig-ms}$ in our framework.  Given a triplet, the self-similarity of the selected positive pair and negative pair are $S_{ap}$ and $S_{an}$. The similarity of other positives and negatives to the anchor is considered as relative-similarity, noted as ${R_{ap}}^{i}$ and ${R_{an}}^{j}$. In addition, \cite{wang2019multi} also defines $\mathcal{P}$ and $\mathcal{N}$ be the sets of selected ${R_{ap}}^{i}$ and ${R_{an}}^{j}$, where 
\begin{align*}
\mathcal{P}=\{{R_{ap}}^{i} \colon {R_{ap}}^{i}<\max\{S_{an}, {R_{an}}^{j}\}+\epsilon\}\\
\mathcal{N}=\{{R_{an}}^{j} \colon {R_{an}}^{j}>\min\{S_{ap}, {R_{ap}}^{i}\}-\epsilon\} \numberthis \label{eq:set}
\end{align*}

\begin{equation}
\left\{
\begin{aligned}
&P_{+}^{sig-ms}=\frac{1}{m_{+}^{sig}+\exp{( \alpha(S_{ap}-\lambda))}}\\
&P_{-}^{sig-ms}=\frac{1}{m_{-}^{sig}+\exp{(-\beta (S_{an}-\lambda))}}
\end{aligned} 
\right.
\label{eq:wp_sigms}
\end{equation}
where 
\begin{align*}
&m_{+}^{sig}=\frac{1}{\left | \mathcal{P} \right |}\sum_\mathcal{P} \exp{( \alpha (S_{ap} - {R_{ap}}^{i}))}\\
&m_{-}^{sig}=\frac{1}{\left | \mathcal{N} \right |}\sum_\mathcal{N} \exp{(-\beta  (S_{an} - {R_{an}}^{j}))}
\end{align*}

When $m_{+}^{sig}=m_{-}^{sig}=1$ the pair weights simplify back to the sigmoid form in equation~\ref{eq:wp_sig}. 

In practice, training MS loss needs to tune four hyper-parameters $\alpha$, $\beta$, $\lambda$ and $\epsilon$ to fit different datasets, making the training not convenient and not efficient. With analysis on relative-similarity terms $m_{+}^{sig}$ and $m_{-}^{sig}$ in the appendix, we define a clearer and parameter free version of pair weight called linear MS pair weight $P^{lin-ms}$, which behaves similar to the original MS weight:
\begin{equation}
\left\{
\begin{aligned}
&P_{+}^{lin-ms}=(1-m_{+}^{lin})(1-S_{ap})\\
&P_{-}^{lin-ms}=(1+m_{-}^{lin})S_{an}
\end{aligned} 
\right.
\label{eq:wp_linms}
\end{equation}where 
\begin{align*}
&m_{+}^{lin}=\frac{1}{\left | \mathcal{P} \right | }\sum_\mathcal{P} (S_{ap}-{R_{ap}}^{i})\\
&m_{-}^{lin}=\frac{1}{\left | \mathcal{N} \right | }\sum_\mathcal{N} (S_{an}-{R_{an}}^{j})
\end{align*}

\subsection{Triplet Weight}
\label{sec:idea_wt}
\begin{figure}[t]
    \centering
    \includegraphics[width=.34\columnwidth]{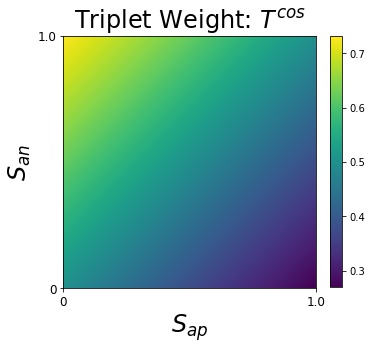}
    \includegraphics[width=.34\columnwidth]{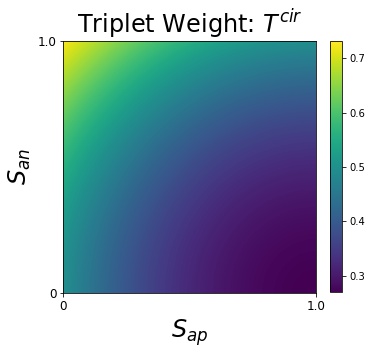}
    \includegraphics[width=.30\columnwidth]{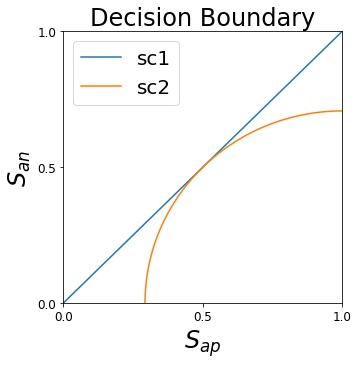}
    \caption{A triplet diagram characterizes the behavior of triplet weights as a function of the the similarity of the anchor-positive pair (along the x-axis) and the anchor-negative pair (along the y-axis).  Triplets where the anchor, positive and negative features are all very similar will be in the top right of the right, and triplets where the anchor-positive are similar and the anchor-negative are not similar are in the bottom right corner.  Using this diagram, (Left) shows the weight of the different triplets based on cosine similarity based triplet loss $T^{cos}$, (Middle) shows the weights of different triplets for the Circle loss $T^{cir}$.  (Right) Shows the decision boundary for two forms of the selectively-contrastive triplet operators $T^{sc1}$ and $T^{sc2}$ (where the anchor-positive pair only has a positive weight below the boundary).
    }
    \label{fig:exp_wt}
\end{figure}
The triplet weight contains the similarity of both positive and negative pairs of a triplet, measuring whether a triplet is well separated or not. In Euclidean distance based triplet loss, the triplet weight (denoted as $T$) is a constant indicating that every triplet will be treated equally. For the fair comparison for other triplet weights, we set constant weight $0.5$. 
\begin{equation}
T^{con}=0.5
\end{equation}

In cosine similarity based triplet loss, the triplet weight is:
\begin{equation}
T^{cos}=\frac{1}{1+\exp{(\tau(S_{ap}-S_{an}))}}  
\label{eq:wt_nca}
\end{equation}
$T^{cos}$ is rely on the difference of $S_{ap}$ and $S_{an}$. When a triplet in a correct configuration, $S_{ap}-S_{an}>0$, the triplet weight is small. Otherwise, the triplet weight will be large. 

In Circle loss~\cite{Sun_2020_CVPR}, the triplet weight is:
\begin{equation}
T^{cir}=\frac{1}{1+\exp{(\tau(S_{ap}(2-S_{ap})-S_{an}^2))}}
\label{eq:wt_cir}
\end{equation}
Because $T^{cos}$ only considers the similarity difference $S_{ap}-S_{an}$, some corner cases such triplet with both large $S_{ap}$ and $S_{an}$ or both small $S_{ap}$ and $S_{an}$ are not well treated.  The idea of $T^{cir}$ is to introduce a non-linear mapping for $S_{ap}$ and $S_{an}$ in the exponential term in order to weight more on the corner cases. 

Figure~\ref{fig:exp_wt} shows the triplet weight diagram, a triplet visualization tool from~\cite{Xuan_2020_ECCV}, for $T^{cos}$ and $T^{cir}$ with $\tau=1$. The equal weight line in $T^{cos}$ is straight lines with form $S_{ap}-S_{an}=\text{const.}$. And the equal weight line in $T^{cir}$ is circular lines with form $(S_{ap}-1)^2+S_{an}^2=\text{const.}$. 

Selectively Contrastive Triplet(SCT) loss~\cite{Xuan_2020_ECCV} selects triplets with hard negatives (the negative example in a triplet is closer to anchor than the positive example) and applies only contrastive loss to the hard negative pairs during the batch training. At gradient level, this approach is to remove the gradients from the anchor-positive pairs for triplets with hard negatives. We treat the selection as a masking operator on positive pair weight:
\begin{eqnarray}
T^{sc1}(P_{+})=
\begin{cases}
0       & \text{if $S_{an}>S_{ap}$}\\
P_{+} & \text{others}
\end{cases}
\label{eq:wt_sc1}
\end{eqnarray}

Because the decision boundary of triplets selection  $S_{an}=S_{ap}$ is a 1st order straight line, we note this masking operator is noted as $T^{sc1}$.  Besides, we continue to extend the selection idea with Circle loss. The triplets in the corner cases can be also selected to only separate the negative pairs. Then, the decision boundary of the selection operator becomes a 2nd order circular line. We note it as $T^{sc2}$, 
\begin{eqnarray}
T^{sc2}(P_{+})=
\begin{cases}
0     & \text{if $S_{ap}(2-S_{ap})-S_{an}^2>0.5$}\\
P_{+} & \text{others}
\end{cases}
\label{eq:wt_sc2}
\end{eqnarray}
Figure~\ref{fig:exp_wt} right shows the difference decision boundaries of $T^{sc1}$ and $T^{sc2}$.

\subsection{Metric Learning Gradient Summary}
\begin{table}[t]
\centering
\begin{adjustbox}{width=1\columnwidth}
\begin{tabular}{l|c|c|c}
Method & Direction & PairWeight & TripletWeight \\
\hline
Triplet (Euclidean)~\cite{facenet}
& $\mathbf{e^{euc}}$      & $P^{euc}$    & $T^{con}$ \\
Triplet (cosine)~\cite{Xuan_2020_WACV}
& $\mathbf{e^{cos}}$      & $P^{con}$    & $T^{cos}$ \\
Circle loss~\cite{Sun_2020_CVPR}
& $\mathbf{e^{cos}}$      & $P^{lin}$    & $T^{cir}$ \\
Binomial deviance~\cite{yi2014deep}
& $\mathbf{e^{cos}}$      & $P^{sig}$    & $T^{con}$ \\
MS loss~\cite{wang2019multi}
& $\mathbf{e^{cos}}$      & $P^{sig-ms}$ & $T^{con}$ \\
DR-MS loss~\cite{Mohan_2020_CVPR}
& $\mathbf{e^{cos-orth}}$ & $P^{sig-ms}$ & $T^{con}$ \\
SC triplet loss~\cite{Xuan_2020_ECCV}
& $\mathbf{e^{cos}}$      & $P^{con}$   & $T^{cos}$, $T^{sc1}$ \\
\hline
\end{tabular}
\end{adjustbox}
\caption{Triplet loss functions define a gradient on the embedded feature locations of the anchor, positive, and negative examples of the triplet.  A large collection of recently proposed triplet loss functions (left) can be put into a unified framework by decomposing the gradient into the (unit) directions they impose on the features, and the weight of that gradient due to the properties of the anchor-positive and anchor-negative pairs, and the overall configuration of the triplet.  This decomposition gives some insight into why some approaches give improved results, and provides a design space for choosing particular combinations of weights to optimize overall performance.}
\label{table:methods}
\end{table}
In this section, we have derived ways to represent many previous loss functions in terms of their gradients.  We have explicitly defined the gradients in terms of how the anchor, positive and negative are moved, defined them in terms of a unit vector in the direction of motion, a weight of anchor-positive term and the anchor negative term and weight of the triplet overall. Table~\ref{table:methods} shows how to map different combinations of gradient components into currently proposed loss functions. Section~\ref{sec:exp_all} gives explicit experiments to understand the isolated effects of these three parts of gradient component.

\section{Experiment Settings}
\label{sec:experiments}
We run a set of experiments on the CUB200 (CUB)~\cite{CUB200}, CAR196 (CAR)~\cite{CAR196}, Stanford Online Products (SOP)~\cite{SOP} and In-shop Cloth (In-shop)~\cite{ICR} dataset. All experiments are run on the PyTorch platform~\cite{pytorch} with Nvidia Tesla V100 GPU, using ResNet~\cite{resnet} architectures, pre-trained on ILSVRC 2012-CLS data~\cite{ILSVRC15}. Training images augmented using a standard scheme (random horizontal flip and random crops padded by 10 pixels on each side), and normalized using the channel means and standard deviations. The network is trained with stochastic gradient descent (SGD) with momentum $0$, step $0.1$ and milestone at $60\%$ of the total epochs. We refer the Easy Positive with Hard Negative mining protocol~\cite{Xuan_2020_WACV} to sample a batch with $C$ classes and $N$ images per class. On CUB, CAR, SOP and In-shop dataset, we sample 8, 16, 4 and 4 images per class in a mini-batch. 

\textbf{Small embedding size comparing to training classes size}: We follow the early goal of deep metric learning works~\cite{SOP,Npairs,Proxy,harwood2017smart} which sets the embedding size to be smaller than the number of training classes. On CUB, CAR, SOP and In-shop dataset the embedding size is 64, 64, 512, 512.

\textbf{Comparison of Gradient Components}: To compare each component in the gradient, we train ResNet18 on CAR dataset and In-shop dataset for 60 epochs. The training is run with batch size 128. For a given test setting, we run the test 5 times to remove the effect caused by the randomness coming from the random sampling of the batch and random initialization of the final FC embedding layer which reducing the GAP feature to a target dimension (e.g. 64 or 512). Then, the mean and standard deviation of Recall@1 are calculated. 

\textbf{Comparison with the State-of-the-Art}: To compare the recent state-of-the-Arts results, we select ResNet50 as the backbone for 80 epochs training. The training is run with different batch sizes 128, 256, 384 and 512. Each test is run 3 times and mean Recall@K is calculated as the measurement for retrieval quality. 

\textbf{PyTorch Implementation}: In PyTorch platform, we use \textit{torch.autograd.Function} module to customize both forward and backward functions for a loss module. The backward function is to generate our customized gradient for the optimizer. During the training, the gradient is directly starting from the backward function, replacing the gradient generated by AutoGrad of the forward function. 

\section{Comparison Experiments}
\label{sec:exp_all}
In this section, we give explicit experiment results to demonstrate the isolated effects contributed by unit gradient direction, pair weight and triplet weight. More raw results are shown in Appendix.

\begin{table}[t]
\centering
\begin{adjustbox}{width=0.8\columnwidth}
\begin{tabular}{c|c|c}
Direction & CAR & In-shop \\
\hline
$\mathbf{e^{euc}}$      & \cellcolor[HTML]{F9DEDC}69.5 $\pm$ 0.7 & \cellcolor[HTML]{FFFFFF}83.7 $\pm$ 0.1 \\
$\mathbf{e^{cos}}$      & \cellcolor[HTML]{EA9088}75.5 $\pm$ 0.2 & \cellcolor[HTML]{F3BCB7}85.2 $\pm$ 0.3 \\
$\mathbf{e^{euc-orth}}$ & \cellcolor[HTML]{FFFFFF}66.9 $\pm$ 0.5 & \cellcolor[HTML]{FCEFEE}84.1 $\pm$ 0.1 \\
$\mathbf{e^{cos-orth}}$ & \cellcolor[HTML]{E67C73}77.0 $\pm$ 0.7 & \cellcolor[HTML]{E67C73}86.6 $\pm$ 0.2
\end{tabular}
\end{adjustbox}
\caption{Comparing recall@1 performance of different gradient directions on CAR and In-shop dataset}
\label{table:grad_gd}
\end{table}
\begin{table}[t]
\centering
\begin{adjustbox}{width=1.0\columnwidth}
\begin{tabular}{c|l|c|c}
 & PairWeight & $\mathbf{e^{euc}}$ & $\mathbf{e^{cos}}$ \\
\hline
 & $P^{con}$ & \cellcolor[HTML]{FFFFFF}69.5 $\pm$ 0.7 & \cellcolor[HTML]{B0DFC8}75.5 $\pm$ 0.2 \\
 & $P^{euc}$ & \cellcolor[HTML]{B4E1CB}75.2 $\pm$ 0.4 & \cellcolor[HTML]{8CD1AF}77.0 $\pm$ 0.4 \\
 & $P^{lin}$ & \cellcolor[HTML]{95D4B5}76.7 $\pm$ 0.5 & \cellcolor[HTML]{75C79F}77.8 $\pm$ 0.9 \\
 & $P^{lin-ms}$ & \cellcolor[HTML]{68C296}78.2 $\pm$ 0.4 & \cellcolor[HTML]{57BB8A}78.8 $\pm$ 0.9 \\
 & $P^{sig}$ & \cellcolor[HTML]{DFF2E9}71.9 $\pm$ 0.4 & \cellcolor[HTML]{C0E6D4}74.3 $\pm$ 0.2 \\
\multirow{-6}{*}{\rotatebox[origin=c]{90}{CAR}}
 & $P^{sig-ms}$ & \cellcolor[HTML]{BCE4D1}74.6 $\pm$ 0.8 & \cellcolor[HTML]{A2DABF}76.2 $\pm$ 0.5 \\
\hline
 & $P^{con}$ & \cellcolor[HTML]{FFFFFF}83.7 $\pm$ 0.1 & \cellcolor[HTML]{CEECDD}85.2 $\pm$ 0.3 \\
 & $P^{euc}$ & \cellcolor[HTML]{D0ECDF}85.2 $\pm$ 0.2 & \cellcolor[HTML]{DBF1E7}84.8 $\pm$ 0.2 \\
 & $P^{lin}$ & \cellcolor[HTML]{6CC499}87.4 $\pm$ 0.1 & \cellcolor[HTML]{70C69C}87.3 $\pm$ 0.2 \\
 & $P^{lin-ms}$ & \cellcolor[HTML]{67C296}87.5 $\pm$ 0.2 & \cellcolor[HTML]{73C79E}87.3 $\pm$ 0.1 \\
 & $P^{sig}$ & \cellcolor[HTML]{AFDFC8}86.2 $\pm$ 0.1 & \cellcolor[HTML]{57BB8A}87.8 $\pm$ 0.2 \\
\multirow{-6}{*}{\rotatebox[origin=c]{90}{In-shop}} & $P^{sig-ms}$ & \cellcolor[HTML]{D9F0E5}84.9 $\pm$ 0.4 & \cellcolor[HTML]{A5DBC1}86.4 $\pm$ 0.2
\end{tabular}
\end{adjustbox}
\caption{Comparing recall@1 performance of different pair weights with Euclidean and cosine direction on CAR and In-shop dataset}
\label{table:grad_wp}
\end{table}
\begin{table}[t]
\centering
\begin{adjustbox}{width=1.0\columnwidth}
\begin{tabular}{c|l|c|c}
 & TripletWeight & $P^{con}$ & $P^{lin}$ \\
\hline
 & \cellcolor[HTML]{FFFFFF}$T^{con}$ & \cellcolor[HTML]{FCFDFF}75.5 $\pm$ 0.2 & \cellcolor[HTML]{7FACF8}77.8 $\pm$ 0.9 \\
 & \cellcolor[HTML]{FFFFFF}$T^{cos}$ & \cellcolor[HTML]{EFF5FF}75.8 $\pm$ 0.2 & \cellcolor[HTML]{8EB6F9}77.5 $\pm$ 0.5 \\
 & \cellcolor[HTML]{FFFFFF}$T^{cos}$ \& $T^{sc1}$ & \cellcolor[HTML]{AAC8FB}77.0 $\pm$ 0.2 & \cellcolor[HTML]{4587F5}78.8 $\pm$ 0.6 \\
 & \cellcolor[HTML]{FFFFFF}$T^{cos}$ \& $T^{sc2}$ & \cellcolor[HTML]{A0C2FA}77.2 $\pm$ 0.5 & \cellcolor[HTML]{5B95F6}78.4 $\pm$ 0.5 \\
 & \cellcolor[HTML]{FFFFFF}$T^{cir}$ & \cellcolor[HTML]{FFFFFF}75.5 $\pm$ 0.3 & \cellcolor[HTML]{6098F6}78.3 $\pm$ 0.2 \\
 & \cellcolor[HTML]{FFFFFF}$T^{cir}$  \& $T^{sc1}$ & \cellcolor[HTML]{A9C8FA}77.0 $\pm$ 0.4 & \cellcolor[HTML]{6EA1F7}78.1 $\pm$ 0.6 \\
\multirow{-7}{*}{\rotatebox[origin=c]{90}{CAR}} & \cellcolor[HTML]{FFFFFF}$T^{cir}$ \& $T^{sc2}$ & \cellcolor[HTML]{BFD6FC}76.6 $\pm$ 0.8 & \cellcolor[HTML]{4285F4}78.8 $\pm$ 0.3 \\
\hline
 & \cellcolor[HTML]{FFFFFF}$T^{con}$ & \cellcolor[HTML]{C9DDFC}85.2 $\pm$ 0.1 & \cellcolor[HTML]{5D97F6}87.3 $\pm$ 0.2 \\
 & \cellcolor[HTML]{FFFFFF}$T^{cos}$ & \cellcolor[HTML]{A4C4FA}86.0 $\pm$ 0.3 & \cellcolor[HTML]{4285F4}87.9 $\pm$ 0.4 \\
 & \cellcolor[HTML]{FFFFFF}$T^{cos}$ \& $T^{sc1}$ & \cellcolor[HTML]{D0E1FD}85.1 $\pm$ 0.3 & \cellcolor[HTML]{5E97F6}87.3 $\pm$ 0.3 \\
 & \cellcolor[HTML]{FFFFFF}$T^{cos}$ \& $T^{sc2}$ & \cellcolor[HTML]{EFF5FF}84.5 $\pm$ 0.2 & \cellcolor[HTML]{76A7F7}86.9 $\pm$ 0.2 \\
 & \cellcolor[HTML]{FFFFFF}$T^{cir}$ & \cellcolor[HTML]{A0C2FA}86.0 $\pm$ 0.1 & \cellcolor[HTML]{4D8CF5}87.7 $\pm$ 0.2 \\
 & \cellcolor[HTML]{FFFFFF}$T^{cir}$  \& $T^{sc1}$ & \cellcolor[HTML]{DCE9FD}84.9 $\pm$ 0.2 & \cellcolor[HTML]{649BF6}87.2 $\pm$ 0.1 \\
\multirow{-7}{*}{\rotatebox[origin=c]{90}{In-shop}} & \cellcolor[HTML]{FFFFFF}$T^{cir}$ \& $T^{sc2}$ & \cellcolor[HTML]{FFFFFF}84.2 $\pm$ 0.1 & \cellcolor[HTML]{73A5F7}86.9 $\pm$ 0.2
\end{tabular}
\end{adjustbox}
\caption{Comparing recall@1 performance of different triplet weights with constant and linear pair weight on CAR and In-shop dataset}
\label{table:grad_wt}
\end{table}
\subsection{Unit Gradient Direction}
\label{sec:exp_gd}
To understand the behavior of unit gradient directions in section~\ref{sec:idea_gd}, we set constant pair and triplet weight $T^{con}=0.5$ and $P^{con}=1$, and vary the choice of  Euclidean, cosine, Euclidean-orthogonal and cosine-orthogonal direction. 

In Table~\ref{table:grad_gd}, we find the following trends.  First, the cosine and cosine-orthogonal direction have better Recall@1 accuracy than other directions.  Second, the cosine-orthogonal gradient direction gives an improvement for both datasets compared to the cos direction. More analysis will be discussed in section~\ref{sec:euc_cos}.

\subsection{Pair Weight}
\label{sec:exp_wp}
\begin{figure}[t]
    \centering
    \includegraphics[width=0.49\columnwidth]{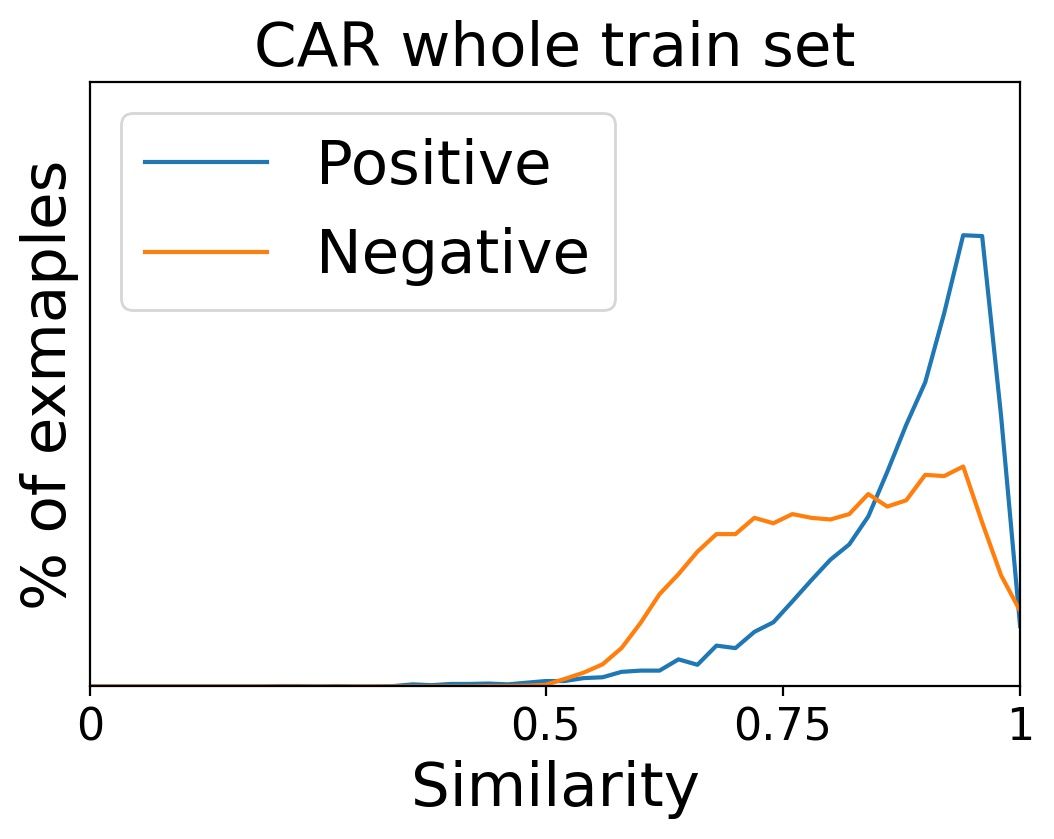}
    \includegraphics[width=0.49\columnwidth]{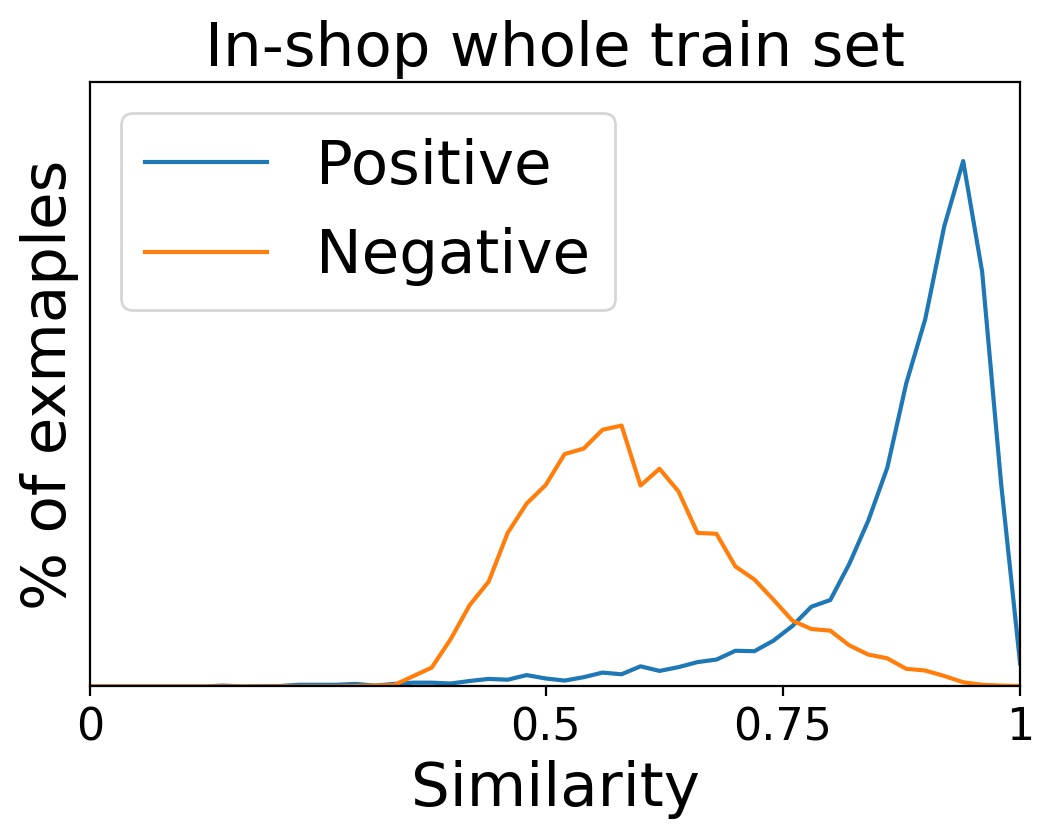}
    \caption{Distribution of the nearest positive pairs(blue) and the nearest negative pairs(orange) over whole CAR(left) and In-shop(right) dataset after training. On the CAR datasets, the nearest positive pairs distribution is largely overlay on the nearest negative pair distribution, indicating the CAR set is not easily to be separated well among different classes. However, on the In-shop datasets, the overlay area of the nearest positive pairs distribution and the nearest negative pair is much less, indicating the In-shop dataset can be easily separated among different classes.
    }
    \label{fig:datasets}
\end{figure}
To understand the behavior of the pair weights, we set triplet weights with constant form $T^{con}=0.5$ and gradient direction  with $\mathbf{e^{cos}}$ and $\mathbf{e^{euc}}$ for two sets of results respectively. As for baseline results, the pair weights are set with constant form $P^{con}=1$. 

In Table~\ref{table:grad_wp}, all pair weights provide a clear performance gain to their baseline results. Also, the performance gap of gradient direction $\mathbf{e^{euc}}$ and $\mathbf{e^{cos}}$ after applying the pair weight is greatly reduced.

Both relative-similarity methods $P^{lin-ms}$ and $P^{sig-ms}$ perform better than the method with only self-similarity on CAR dataset across different learning rates. Due to the property of well separation on In-shop dataset as shown in Figure~\ref{fig:datasets}, the relative-similarity term will less likely exist during the train because few positive and negative examples will be in $\mathcal{P}$ and $\mathcal{N}$ set as mentioned in equation~\ref{eq:set}. $P^{lin-ms}$ is performance almost as same as $P^{lin}$, but $P^{sig-ms}$ shows some computation instability effect. We put a further analysis of this effect in the appendix.

In summary, Table~\ref{table:grad_wp} shows several features related to the pair weight.  First, pair weight causes substantial improvement in recall@1 accuracy. Second, in most cases, the linear and sigmoid pair weight outperforms the default Euclidean pair weight. Third, the linear version of the multi-similarity gradient direction is much more robust to different learning rates than the sigmoid version(see in appendix), and gives better performance and Recall@1 accuracy. 

\subsection{Triplet Weight}
\label{sec:exp_wt}
Table~\ref{table:grad_wt}, we show two groups of experiments to compare the seven triplet weights. One group sets the pair-weight to be constant $P^{con}=1$.  Another group uses the linear pair weight $P^{lin}$. All experiments use cosine gradient direction. 

Comparing to the baseline method where triplet weight $T^{con}=0.5$, $T^{cos}$ and $T^{cir}$ has minimal but slight boost in performance; $T^{sc1}$ and $T^{sc2}$ has bigger impact on CAR data set than In-shop dataset. This is due to the properties of these two datasets as shown in Figure~\ref{fig:datasets}. CAR dataset has low inter-class variance(images from different classes may look similar) while In-shop dataset has high inter-class variance(images from different classes look not similar). The major challenge of CAR dataset is to distinguish similar images with different labels, and this is the purpose of triplet operators $T^{sc1}$ and $T^{sc2}$ because they concentrate on separating triplets with hard negative in training.  And In-shop dataset is to relatively easy to separate images with different labels, the goal is to continue separating the images better, which is the impact of $T^{cos}$ and $T^{cir}$

Therefore, we can conclude that the performance gain in Circle loss is largely from the pair weight not from the triplet weight. Selective Contrastive operator benefits the training tasks which need to separate triplets with hard negative and is not helpful for training tasks which easily separate triplets during the training. 


\subsection{Euclidean or Cosine Direction?}
\label{sec:euc_cos}
\begin{table*}[t]
\centering
\begin{adjustbox}{width=1.0\textwidth}
\begin{tabular}{l|ccc|ccc|ccc|ccc}
Dataset & \multicolumn{3}{c}{CUB(dim=64)} & \multicolumn{3}{c}{CAR(dim=64)} & \multicolumn{3}{c}{SOP(dim=512)} & \multicolumn{3}{c}{In-shop(dim=512)} \\
\hline
Method & R@1 & R@2 & R@4 & R@1 & R@2 & R@4 & R@1 & R@10 & R@100 & R@1 & R@10 & R@20 \\
\hline
LiftedStruct~\cite{SOP} 
& 43.6 & 56.6 & 68.6 & 53.0 & 65.7 & 76.0 & 62.5 & 80.8 & 91.9 & - & - & - \\
ProxyNCA~\cite{Proxy} 
& 49.2 & 61.9 & 67.9 & 73.2 & 82.4 & 86.4 & 73.7 & - & - & - & - & - \\
SoftTriple~\cite{Qian_2019_ICCV} 
& 60.1 & 71.9 & 81.2 & 78.6 & 86.6 & 91.8 & 78.3 & 90.3 & 95.9 & - & - & - \\
EasyPositive~\cite{Xuan_2020_WACV} 
& 57.3 & 68.9 & 79.3 & 75.5 & 84.2 & 90.3 & 78.3 & 90.7 & 96.3 & 87.8 & 95.7 & 96.8 \\
MS~\cite{wang2019multi} 
& 57.4 & 69.8 & 80.0 & 77.3 & 85.3 & 90.5 & 78.2 & 90.5 & 96.0 & 89.7 & 97.9 & 98.5 \\
SCT~\cite{Xuan_2020_ECCV}
& 57.7 & 69.8 & 79.6 & 73.4 & 82.0 & 88.0 & 81.9 & 92.6 & 96.8 & 90.9 & 97.5 & 98.2 \\
DR-MS~\cite{Mohan_2020_CVPR} 
& 59.1 & 71.0 & 80.3 & 79.3 & 86.7 & 91.4 & - & - & - & 91.7 & \bf{98.1} & 98.7 \\
Proxy-anchor~\cite{Kim_2020_CVPR} & 61.7 & 73.0 & 81.8 & 78.8 & 87.0 & 92.2 & 79.1 & 90.8 & 96.2 & 91.5 & \bf{98.1} & \bf{98.8} \\
\hline
MS*(B128)
& 59.8 & 71.7 & 81.0 & 79.0 & 86.6 & 91.5 & 78.7 & 90.4 & 96.0 & 89.4 & 96.6 & 97.4 \\
DR-MS*(B128)
& 60.7 & 71.9 & 81.3 & 79.9 & 87.0 & 91.7 & 78.8 & 90.4 & 96.1 & 89.6 & 96.4 & 97.4\\
Ours(B128) 
& 63.5 & 74.8 & 83.6 & 82.5 & 89.1 & 93.3 & 79.9 & 90.5 & 95.5 & 91.4 & 97.7 & 98.4 \\
Ours(B256) 
& \bf{63.8} & 74.8 & 83.7 & 85.5 & 91.0 & 94.6 & 82.0 & 92.3 & 96.8 & \bf{92.2} & 97.8 & 98.4 \\
Ours(B384) 
& \bf{63.8} & \bf{75.0} & \bf{84.2} & \bf{86.5} & \bf{91.6} & \bf{94.8} & 82.2 & \bf{92.5} & 96.8 & 92.0 & 97.8 & 98.3 \\
Ours(B512) 
& 63.1 & 74.6 & 83.2 & 85.7 & 91.2 & 94.7 & \bf{82.3} & \bf{92.5} & \bf{96.9} & 90.8 & 97.2 & 97.9 \\
\hline
\end{tabular}
\end{adjustbox}
\caption{Retrieval Performance on the CUB, CAR, SOP and In-shop datasets comparing to the best reported results.}
\label{table:SOTA}
\end{table*}
\begin{figure}[t]
    \centering
    \includegraphics[width=.32\columnwidth]{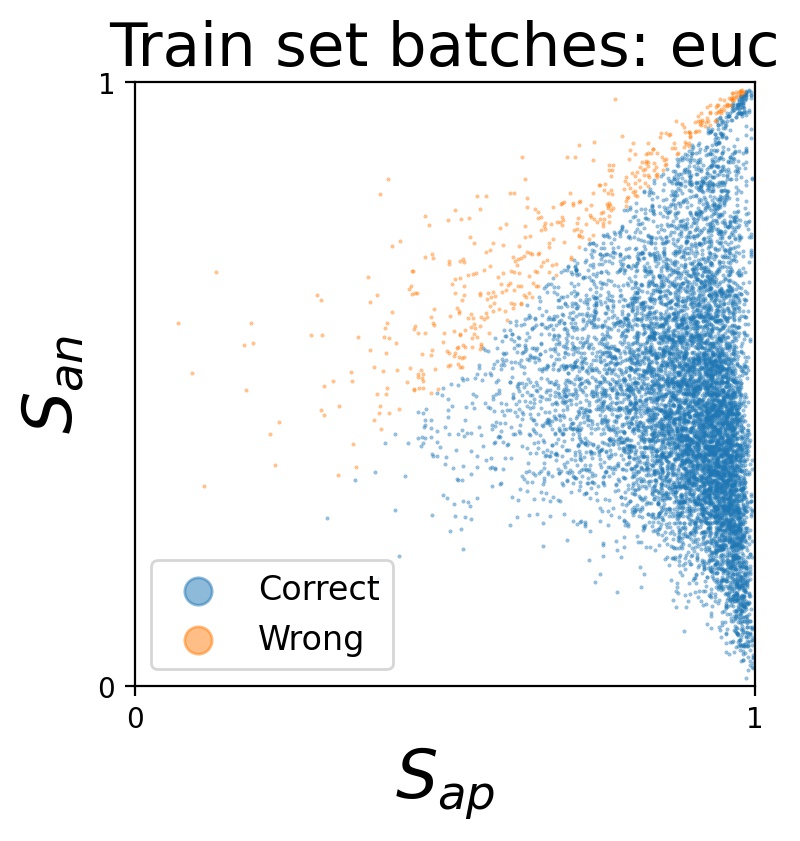}
    \includegraphics[width=.32\columnwidth]{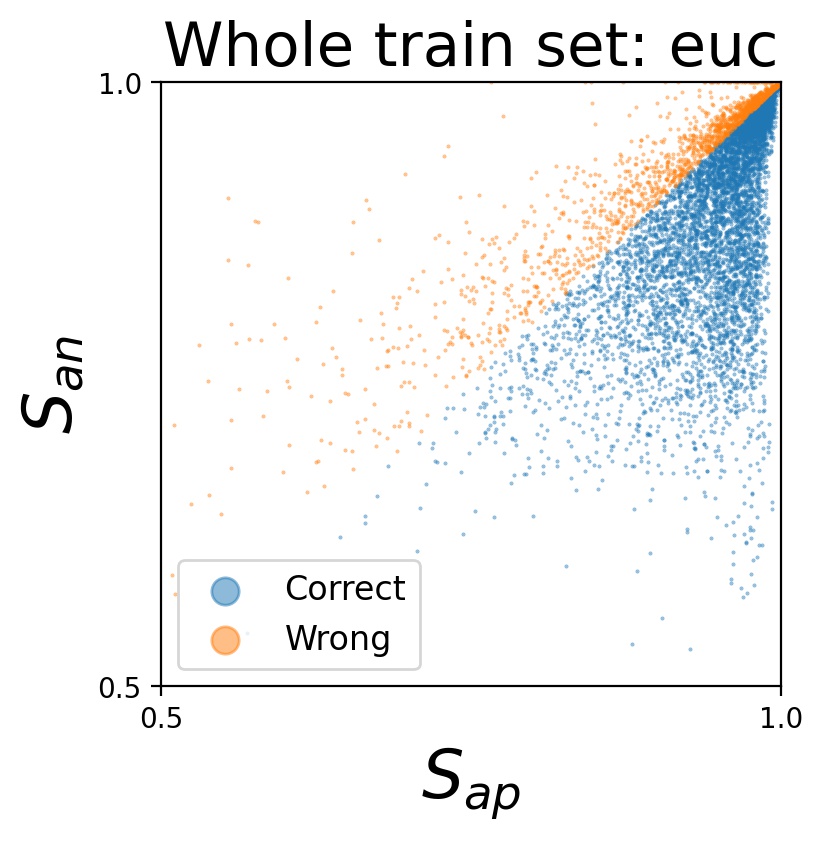}
    \includegraphics[width=.32\columnwidth]{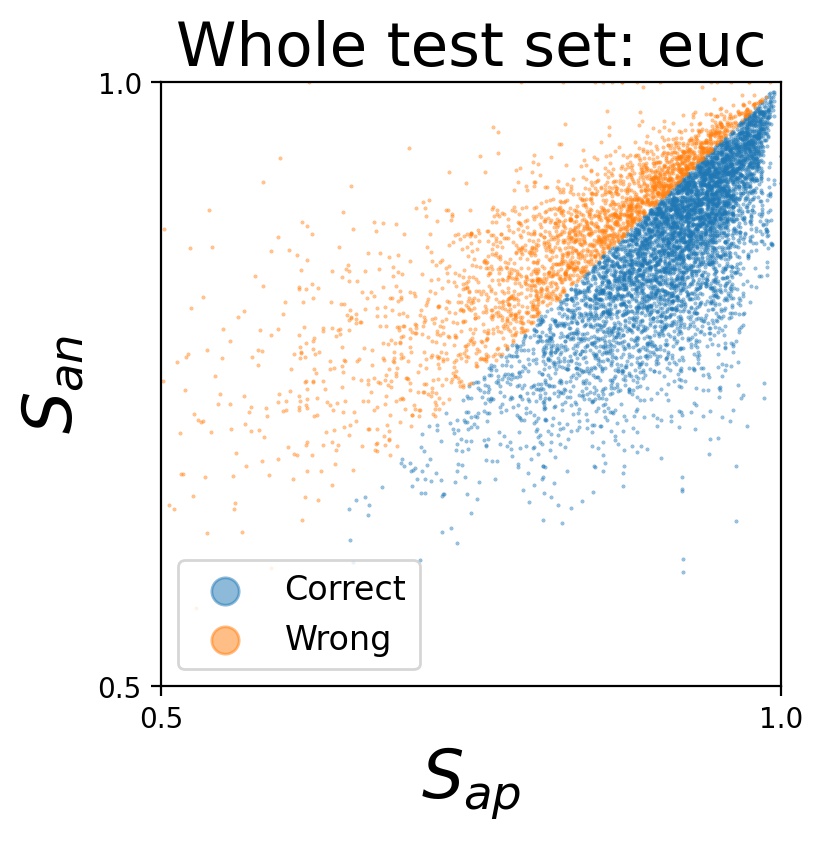}
    \includegraphics[width=.32\columnwidth]{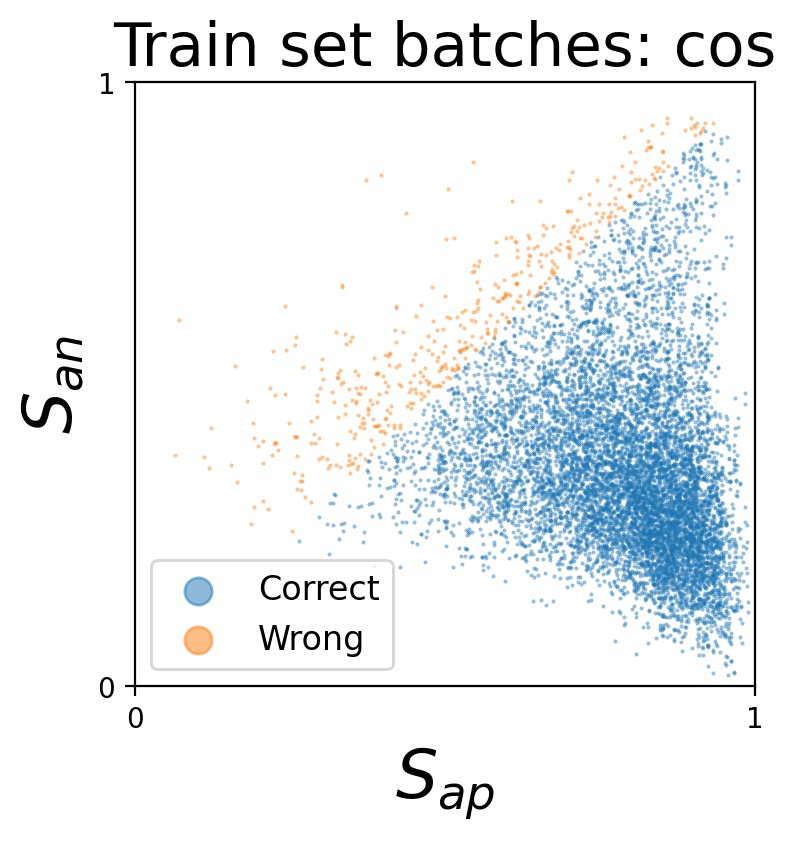}
    \includegraphics[width=.32\columnwidth]{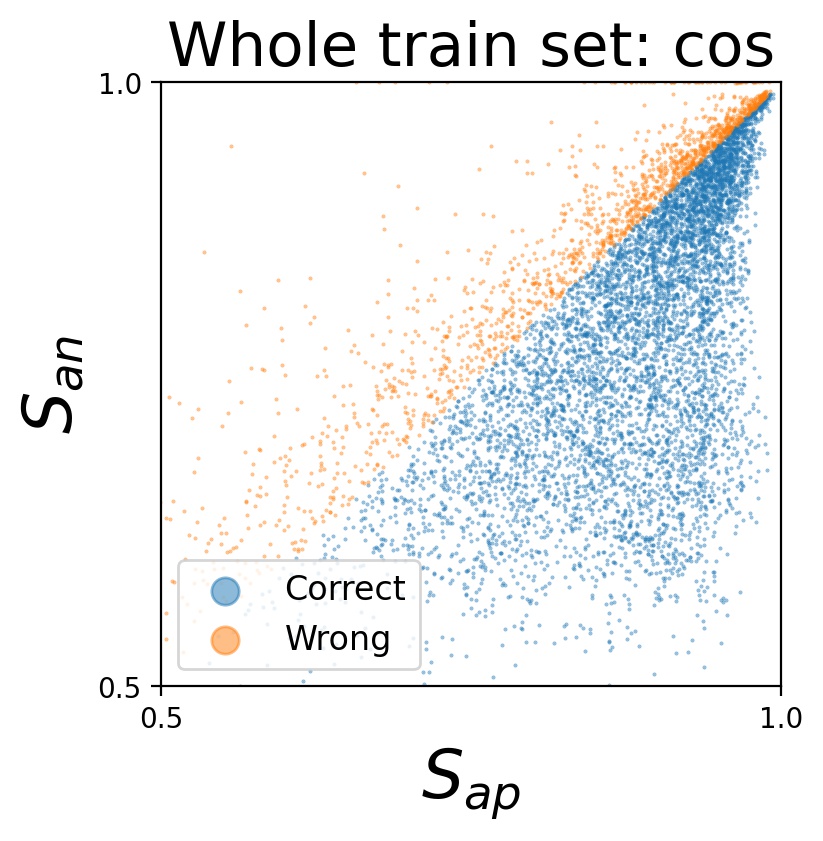}
    \includegraphics[width=.32\columnwidth]{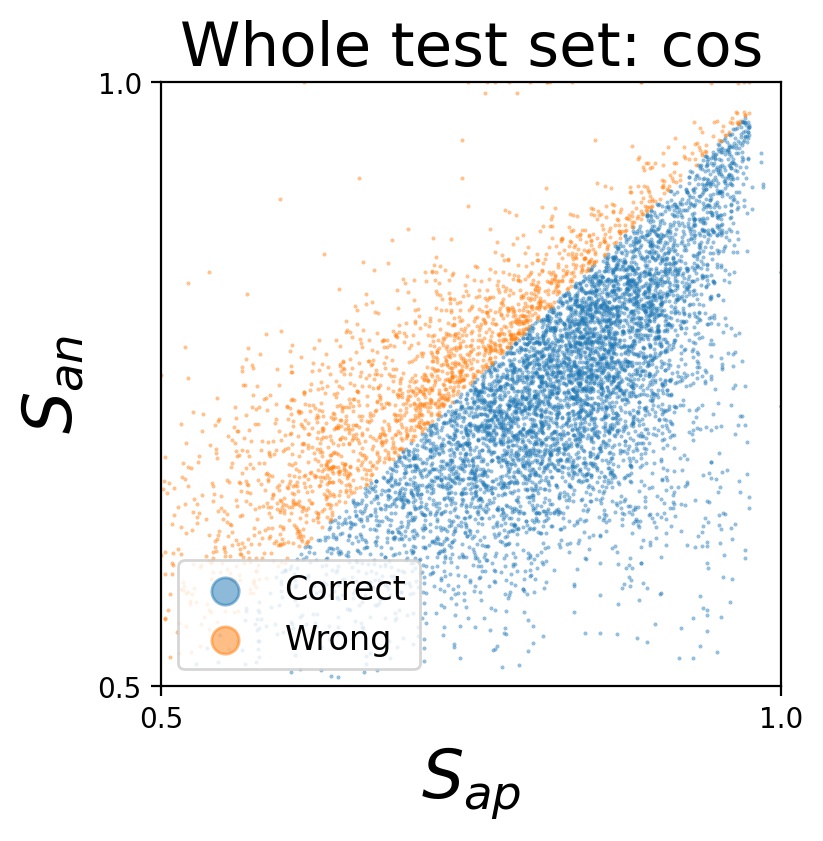}
    \caption{Visualization of triplets mined with the nearest positive and the nearest negative strategy~\cite{Xuan_2020_WACV} in training batches of the last 5 epochs (left), in the whole train set (middle) and in the whole test set (right) after training. First row: Euclidean gradient direction. Second row: Cosine gradient direction.}
    \label{fig:cos_euc_scatter}
\end{figure}

In Section~\ref{sec:idea_gd} and Figure~\ref{fig:cos_euc_thoery}, the different gradient behaviors of $\mathbf{e^{euc}}$ and $\mathbf{e^{cos}}$ have been showed.  But additional discussion 
will highlight the performance difference shown in Sections~\ref{sec:exp_gd} and~\ref{sec:exp_wp}.

We first decompose the unit gradient to move positive and negative features into two directions: the direction along positive and negative feature $d_{\parallel}$ and the direction orthogonal to positive and negative feature $d_{\bot}$. As shown in Figure~\ref{fig:cos_euc_thoery}, only the gradient component along $d_{\bot}$ effectively contributes to the angle change of anchor-positive and anchor-negative pair which directly affect the similarity score.  The effective gradient projection strength for $\mathbf{e^{euc}}$ and $\mathbf{e^{cos}}$:

\begin{eqnarray}
\begin{cases}
\sqrt{\frac{1+S}{2}} & \text{Euclidean direction}\\
\sqrt{1-S^2}         & \text{cosine direction}
\end{cases}
\label{eq:euc_cos_projection}
\end{eqnarray} where $S$ is the similarity of a positive or negative pair. The derivation of the above projection length is shown in the appendix.

Figure~\ref{fig:cos_euc_thoery} right shows the change of the effective gradient strength for $\mathbf{e^{euc}}$ and $\mathbf{e^{cos}}$ varying as a function of pair similarity. Because most pairs during the training have positive similarity, we focus on projection length when similarity is positive. 

The Euclidean gradient has stronger force to pull positive close and push negative away than the cosine direction when two features are close to each other. Therefore, Euclidean gradient continues to force features together even when they are already relatively close, unlike the cosine gradient. In Figure~\ref{fig:cos_euc_scatter} left column, we show the triplet diagram plot of triplets extracted from the last 5 epochs of training (epoch 55-60) on CAR dataset. The Euclidean direction clusters the same label feature more tightly than the cosine direction because there are more triplets along the right edge of the triplet diagram comparing to the scatter of cosine direction. 

However, the tight clustering behavior in training leads to even the triplet with nearest positive and the nearest negative to be compact. In the middle and right column of Figure~\ref{fig:cos_euc_scatter}, we plot just these triplets for the whole training set (middle), and testing set (right). The Euclidean gradient has more triplets very close to the top right corner, indicating that point have very similar same class {\em and} different class neighbors, while cosine gradient creates triplets that are more spread out. The spread out effect indicates the feature learned by the deep model is distinguishable~\cite{Zhang_2017_ICCV,Wu_2018_CVPR,Xuan_2020_WACV}. Because these Euclidean feature are more compressed (for both the anchor-positive and anchor-negative pairs), it is harder for the network to learn distinguishable features that if it is using the cosine gradient.

One more piece of evidence to support the analysis above is the pair weight result in section~\ref{sec:exp_wp}. When the Euclidean pair weight $P^{euc}$ is applied to Euclidean direction and cosine direction, the performance gap between these two methods is almost disappeared. This is because the Euclidean pair weight $P^{euc}$ reduces the weight de-emphasizes positive pair when they are already close, and therefore avoid its tight clustering behavior, making Euclidean direction behave similarly as cosine direction.

\section{Best combination of gradients}

In the previous chapter, we separately consider the gradient terms that relate to the gradient directions, the pair weights applied to the gradients from the anchor-negative and anchor-positive pairs, and the overall weight of the triplets.  In terms of the gradient direction, the $\mathbf{e}^{cos-orth}$ gives the best performance and is relatively stable with respect to the learning rate.  In terms of the pair-weighting, $P^{lin-ms}$ is consistently a top performer across datasets.  Similarly, $T^{cir}$ shows stable improvement to both CAR and In-shop datasets. We combine these gradient components empirically to form the final gradient, and train a network by imposing this gradient combination.  We compare the performance of the network trained this way with many latest state-of-the-art results. 

To ensure a fair comparison, we also re-implement current related SOTA approaches, MS and DR-MS results (noted as MS* and DR-MS*) with our gradient method to create a comparison with the same network backbone, pre-processing and training settings. The implementation difference is shown in the appendix. The result is reported in Table~\ref{table:SOTA}.  In addition, we vary the batch size 128, 256, 384, 512 on all tests for four datasets and continue to improve the Recall performance. 

\section{Limitations}
We point out the following limitations of the paper:
\begin{itemize}
    \item We do not exhaustively compute all possible combination of all the three gradient components, and instead focus on the isolated effect of single gradient components.  There may be additional improvements in explicitly considering the interactions between the different gradient components.
    \item  There are recent loss functions proposed in the deep metric learning literature such as Proxy loss~\cite{Proxy} and N-pair loss~\cite{Npairs}; and we currently are not able to put those loss function into our framework due to complex gradient computation for multiple negative pairs. 
    \item Our experiments do not fully explore training optimizations.  We have fixed hyper-parameters in our sampling approach, we keep a constant step size, and we fix the hyper-parameters in gradient components such as $P^{sig-ms}$ for most experiments. Our results are based on hyper-parameter selections from earlier papers, but the gradient based approach to learning embedding functions may be improved with additional search over the hyper-parameter space.
\end{itemize}

\section{Conclusion}
We provide a new framework to train deep metric learning networks with direct gradient modification. In our framework, we disentangled gradient components of many loss functions into common components, and analyze the effects of each component.  We find that the Euclidean gradient direction and the cosine gradient direction behave quite differently. In its default form, the Euclidean gradient creates embedding spaces that are very tightly clustered and the cosine gradient direction has a consistently big improvement over a large set of experimental conditions. 

Second, recently popular works define new loss functions that, in terms of their gradient, primarily change the pair weight term, which is consistent with our findings that the pair-weight term is very important.  In contrast, we find the triplet weight term to have limited impact that was not consistent across datasets.  

Finally, this study of the importance of different weighting functions and components of the gradient led to a simple approach that directly defines the desired gradients and gives improvements to state-of-the-art performance relative to recent work.

\end{document}


\title{Appendix}

\maketitle
\begin{table*}[t]
\centering
\begin{adjustbox}{width=1.0\textwidth}
\begin{tabular}{c|cccc|cccc}
Dataset & \multicolumn{4}{c|}{CAR} & \multicolumn{4}{c}{In-shop} \\
\hline
Direction & lr 0.025 & lr 0.05 & lr 0.1 & lr 0.2 & lr 0.5 & lr 1.0 & lr 2.0 & lr 4.0 \\
\hline
$\mathbf{e^{euc}}$ & \cellcolor[HTML]{F6CBC6}66.8 $\pm$ 0.4 & \cellcolor[HTML]{F2B8B2}69.5 $\pm$ 0.7 & \cellcolor[HTML]{F6CAC6}66.9 $\pm$ 0.6 & \cellcolor[HTML]{FFFEFD}59.2 $\pm$ 0.2 & \cellcolor[HTML]{FAE2E0}83.7 $\pm$ 0.1 & \cellcolor[HTML]{FAE3E1}83.7 $\pm$ 0.2 & \cellcolor[HTML]{FBE9E7}83.3 $\pm$ 0.2 & \cellcolor[HTML]{FFFFFF}82.0 $\pm$ 0.2 \\
$\mathbf{e^{cos}}$ & \cellcolor[HTML]{F4C3BE}67.9 $\pm$ 0.4 & \cellcolor[HTML]{EB928A}74.3 $\pm$ 0.7 & \cellcolor[HTML]{E98880}75.5 $\pm$ 0.2 & \cellcolor[HTML]{F0ACA6}71.0 $\pm$ 0.5 & \cellcolor[HTML]{F5C8C3}85.2 $\pm$ 0.3 & \cellcolor[HTML]{F6CDC9}84.9 $\pm$ 0.2 & \cellcolor[HTML]{F8D7D4}84.3 $\pm$ 0.4 & \cellcolor[HTML]{FBEAE8}83.3 $\pm$ 0.3 \\
$\mathbf{e^{euc-orth}}$ & \cellcolor[HTML]{FFFFFF}58.9 $\pm$ 0.5 & \cellcolor[HTML]{F6CAC6}66.9 $\pm$ 0.5 & \cellcolor[HTML]{F8D7D3}65.0 $\pm$ 0.6 & \cellcolor[HTML]{FEF9F8}60.0 $\pm$ 0.9 & \cellcolor[HTML]{F9DDDA}84.0 $\pm$ 0.2 & \cellcolor[HTML]{F9DCD9}84.1 $\pm$ 0.1 & \cellcolor[HTML]{FAE2DF}83.7 $\pm$ 0.2 & \cellcolor[HTML]{FEF8F7}82.5 $\pm$ 0.3 \\
$\mathbf{e^{cos-orth}}$ & \cellcolor[HTML]{F2B9B4}69.3 $\pm$ 0.4 & \cellcolor[HTML]{E98B83}75.2 $\pm$ 0.3 & \cellcolor[HTML]{E67C73}77.0 $\pm$ 0.7 & \cellcolor[HTML]{EC978F}73.7 $\pm$ 0.7 & \cellcolor[HTML]{EFA7A1}86.1 $\pm$ 0.2 & \cellcolor[HTML]{EB958D}86.3 $\pm$ 0.4 & \cellcolor[HTML]{E67C73}86.6 $\pm$ 0.2 & \cellcolor[HTML]{F5C7C3}85.3 $\pm$ 0.1 \\
\hline
\end{tabular}
\end{adjustbox}
\caption{Comparing recall@1 performance of different gradient
directions on CAR and In-shop dataset with various learning rates(lr).}
\label{table:grad_gd_raw}
\end{table*}
\begin{table*}[t]
\centering
\begin{adjustbox}{width=1.0\textwidth}
\begin{tabular}{c|c|cccc|cccc}
 \multicolumn{2}{c|}{Dataset} & \multicolumn{4}{c|}{CAR} & \multicolumn{4}{c}{In-shop} \\
 \hline
PairWeight & Direction & lr 0.05 & lr 0.1 & lr 0.2 & lr 0.4 & lr 0.5 & lr 1.0 & lr 2.0 & lr 4.0 \\
\hline
$P^{con}$    & $\mathbf{e^{cos}}$ & \cellcolor[HTML]{A9DDC4}74.3 $\pm$ 0.7 & \cellcolor[HTML]{92D3B4}75.5 $\pm$ 0.2 & \cellcolor[HTML]{C2E7D5}71.0 $\pm$ 0.5 & \cellcolor[HTML]{FFFFFF}62.5 $\pm$ 0.5 & \cellcolor[HTML]{AEDFC7}85.2 $\pm$ 0.3 & \cellcolor[HTML]{AFDFC8}84.9 $\pm$ 0.2 & \cellcolor[HTML]{B1E0C9}84.3 $\pm$ 0.4 & \cellcolor[HTML]{B4E1CB}83.3 $\pm$ 0.3 \\
$P^{euc}$    & $\mathbf{e^{cos}}$ & \cellcolor[HTML]{C0E6D4}71.2 $\pm$ 0.4 & \cellcolor[HTML]{7ECBA6}76.6 $\pm$ 0.5 & \cellcolor[HTML]{78C9A2}77.0 $\pm$ 0.4 & \cellcolor[HTML]{BDE4D1}71.8 $\pm$ 0.7 & \cellcolor[HTML]{AFDFC8}84.8 $\pm$ 0.2 & \cellcolor[HTML]{B0DFC8}84.6 $\pm$ 0.4 & \cellcolor[HTML]{B2E0CA}84.0 $\pm$ 0.2 & \cellcolor[HTML]{B3E1CB}83.3 $\pm$ 0.6 \\
$P^{lin}$    & $\mathbf{e^{cos}}$ & \cellcolor[HTML]{DDF2E8}67.2 $\pm$ 0.6 & \cellcolor[HTML]{AFDFC8}73.7 $\pm$ 0.3 & \cellcolor[HTML]{75C79F}77.2 $\pm$ 0.5 & \cellcolor[HTML]{6AC397}77.8 $\pm$ 0.9 & \cellcolor[HTML]{A8DCC3}86.2 $\pm$ 0.1 & \cellcolor[HTML]{75C89F}87.2 $\pm$ 0.3 & \cellcolor[HTML]{6DC49A}87.3 $\pm$ 0.2 & \cellcolor[HTML]{98D6B8}86.5 $\pm$ 0.4 \\
$P^{lin-ms}$ & $\mathbf{e^{cos}}$ & \cellcolor[HTML]{CBEADB}69.8 $\pm$ 0.9 & \cellcolor[HTML]{8ED2B1}75.8 $\pm$ 0.3 & \cellcolor[HTML]{63C093}78.2 $\pm$ 0.6 & \cellcolor[HTML]{57BB8A}78.8 $\pm$ 0.9 & \cellcolor[HTML]{95D4B6}86.5 $\pm$ 0.1 & \cellcolor[HTML]{70C59C}87.3 $\pm$ 0.1 & \cellcolor[HTML]{7FCBA6}87.0 $\pm$ 0.1 & \cellcolor[HTML]{AEDFC8}85.1 $\pm$ 0.5 \\
$P^{sig}$    & $\mathbf{e^{cos}}$ & \cellcolor[HTML]{DEF2E8}67.2 $\pm$ 0.6 & \cellcolor[HTML]{B5E1CC}72.8 $\pm$ 0.2 & \cellcolor[HTML]{A9DCC4}74.3 $\pm$ 0.2 & \cellcolor[HTML]{ACDEC6}74.0 $\pm$ 0.3 & \cellcolor[HTML]{92D3B4}86.6 $\pm$ 0.1 & \cellcolor[HTML]{6DC49A}87.3 $\pm$ 0.1 & \cellcolor[HTML]{57BB8A}87.8 $\pm$ 0.2 & \cellcolor[HTML]{75C79F}87.2 $\pm$ 0.2 \\
$P^{sig-ms}$ & $\mathbf{e^{cos}}$ & \cellcolor[HTML]{D8EFE4}68.0 $\pm$ 0.6 & \cellcolor[HTML]{B4E1CB}73.0 $\pm$ 0.1 & \cellcolor[HTML]{87CFAC}76.2 $\pm$ 0.5 & \cellcolor[HTML]{89D0AE}76.0 $\pm$ 0.7 & \cellcolor[HTML]{9CD7BB}86.4 $\pm$ 0.2 & \cellcolor[HTML]{ACDEC6}86.0 $\pm$ 0.4 & \cellcolor[HTML]{D3EDE1}72.5 $\pm$ 0.9 & \cellcolor[HTML]{FFFFFF}57.0 $\pm$ 4.5 \\
\hline
$P$ & $\mathbf{e}$ & lr 0.05 & lr 0.1 & lr 0.2 & lr 0.4 & lr 0.5 & lr 1.0 & lr 2.0 & lr 4.0 \\
\hline
$P^{con}$    & $\mathbf{e^{euc}}$ & \cellcolor[HTML]{B7E2CE}69.5 $\pm$ 0.7 & \cellcolor[HTML]{C2E7D5}66.9 $\pm$ 0.6 & \cellcolor[HTML]{E3F4EC}59.2 $\pm$ 0.8 & \cellcolor[HTML]{FFFFFF}52.4 $\pm$ 0.6 & \cellcolor[HTML]{AEDEC7}83.7 $\pm$ 0.1 & \cellcolor[HTML]{AEDEC7}83.7 $\pm$ 0.2 & \cellcolor[HTML]{AEDFC8}83.3 $\pm$ 0.2 & \cellcolor[HTML]{B1E0C9}82.0 $\pm$ 0.2 \\
$P^{euc}$    & $\mathbf{e^{euc}}$ & \cellcolor[HTML]{A7DCC2}72.6 $\pm$ 0.9 & \cellcolor[HTML]{82CDA8}75.2 $\pm$ 0.4 & \cellcolor[HTML]{8FD2B1}74.3 $\pm$ 0.8 & \cellcolor[HTML]{C1E6D4}67.2 $\pm$ 0.5 & \cellcolor[HTML]{ACDEC6}84.9 $\pm$ 0.2 & \cellcolor[HTML]{A8DCC3}85.2 $\pm$ 0.2 & \cellcolor[HTML]{ACDEC6}84.9 $\pm$ 0.3 & \cellcolor[HTML]{AEDEC7}83.7 $\pm$ 0.3 \\
$P^{lin}$    & $\mathbf{e^{euc}}$ & \cellcolor[HTML]{C0E6D3}67.5 $\pm$ 0.8 & \cellcolor[HTML]{9BD7BA}73.4 $\pm$ 0.7 & \cellcolor[HTML]{6EC59B}76.6 $\pm$ 0.7 & \cellcolor[HTML]{6DC49A}76.7 $\pm$ 0.5 & \cellcolor[HTML]{7ECBA6}86.4 $\pm$ 0.1 & \cellcolor[HTML]{5BBD8D}87.4 $\pm$ 0.1 & \cellcolor[HTML]{5BBD8D}87.4 $\pm$ 0.3 & \cellcolor[HTML]{7CCAA4}86.5 $\pm$ 0.4 \\
$P^{lin-ms}$ & $\mathbf{e^{euc}}$ & \cellcolor[HTML]{B6E2CD}69.7 $\pm$ 0.3 & \cellcolor[HTML]{7DCBA5}75.5 $\pm$ 0.6 & \cellcolor[HTML]{5BBD8D}77.9 $\pm$ 0.4 & \cellcolor[HTML]{57BB8A}78.2 $\pm$ 0.4 & \cellcolor[HTML]{7DCBA5}86.4 $\pm$ 0.2 & \cellcolor[HTML]{58BC8B}87.5 $\pm$ 0.2 & \cellcolor[HTML]{57BB8A}87.5 $\pm$ 0.1 & \cellcolor[HTML]{ACDEC6}85.0 $\pm$ 0.3 \\
$P^{sig}$    & $\mathbf{e^{euc}}$ & \cellcolor[HTML]{BEE5D2}68.0 $\pm$ 0.2 & \cellcolor[HTML]{ADDEC6}71.9 $\pm$ 0.4 & \cellcolor[HTML]{B0DFC9}71.1 $\pm$ 0.3 & \cellcolor[HTML]{B8E3CE}69.2 $\pm$ 0.7 & \cellcolor[HTML]{9CD7BB}85.5 $\pm$ 0.2 & \cellcolor[HTML]{90D2B2}85.9 $\pm$ 0.2 & \cellcolor[HTML]{86CEAB}86.2 $\pm$ 0.1 & \cellcolor[HTML]{9BD7BA}85.6 $\pm$ 0.2 \\
$P^{sig-ms}$ & $\mathbf{e^{euc}}$ & \cellcolor[HTML]{BFE5D3}67.7 $\pm$ 1.0 & \cellcolor[HTML]{9CD7BB}73.3 $\pm$ 0.5 & \cellcolor[HTML]{8AD0AE}74.6 $\pm$ 0.8 & \cellcolor[HTML]{93D4B5}74.0 $\pm$ 0.8 & \cellcolor[HTML]{ACDEC6}84.9 $\pm$ 0.4 & \cellcolor[HTML]{C1E6D4}72.4 $\pm$ 1.3 & \cellcolor[HTML]{EAF7F0}48.6 $\pm$ 3.2 & \cellcolor[HTML]{FFFFFF}35.8 $\pm$ 4.9 \\
\hline
\end{tabular}
\end{adjustbox}
\caption{Comparing recall@1 performance of different pair weights on CAR and In-shop dataset with various learning rates(lr).}
\label{table:grad_wp_raw}
\end{table*}
\begin{table*}[t]
\centering
\begin{adjustbox}{width=1.0\textwidth}
\begin{tabular}{c|c|cccc|cccc}
 \multicolumn{2}{c|}{Dataset}& \multicolumn{4}{c|}{CAR} & \multicolumn{4}{c}{In-shop} \\
 \hline
TripletWeight & PairWeight & lr0.05 & lr 0.1 & lr 0.2 & lr 0.4 & lr 0.5 & lr 1.0 & lr 2.0 & lr 4.0 \\
\hline
$T^{con}$ & $P^{con}$ & 
\cellcolor[HTML]{A0C1FA}74.3 $\pm$ 0.7 & \cellcolor[HTML]{77A7F8}75.5 $\pm$ 0.2 & \cellcolor[HTML]{BBD3FC}71.0 $\pm$ 0.5 & \cellcolor[HTML]{FFFFFF}62.5 $\pm$ 0.5 & \cellcolor[HTML]{76A6F8}85.2 $\pm$ 0.1 & \cellcolor[HTML]{88B2F9}84.9 $\pm$ 0.3 & \cellcolor[HTML]{A9C8FB}84.3 $\pm$ 0.2 & \cellcolor[HTML]{D5E4FD}83.3 $\pm$ 0.4 \\
$T^{cos}$ & $P^{con}$ & 
\cellcolor[HTML]{A7C6FB}73.5 $\pm$ 0.3 & \cellcolor[HTML]{70A3F7}75.8 $\pm$ 0.2 & \cellcolor[HTML]{A2C3FB}74.1 $\pm$ 0.6 & \cellcolor[HTML]{DBE8FE}67.0 $\pm$ 0.5 & \cellcolor[HTML]{4C8CF5}85.9 $\pm$ 0.2 & \cellcolor[HTML]{4789F5}86.0 $\pm$ 0.3 & \cellcolor[HTML]{5994F6}85.7 $\pm$ 0.1 & \cellcolor[HTML]{72A4F8}85.3 $\pm$ 0.3 \\
$T^{cos}$,$T^{sc1}$ & $P^{con}$ &  \cellcolor[HTML]{93B9FA}74.7 $\pm$ 0.3 & \cellcolor[HTML]{4889F5}77.0 $\pm$ 0.2 & \cellcolor[HTML]{6CA0F7}75.9 $\pm$ 0.7 & \cellcolor[HTML]{CADDFD}69.2 $\pm$ 1.1 & \cellcolor[HTML]{86B1F9}85.0 $\pm$ 0.2 & \cellcolor[HTML]{7EACF8}85.1 $\pm$ 0.3 & \cellcolor[HTML]{9FC1FA}84.6 $\pm$ 0.1 & \cellcolor[HTML]{EDF4FF}82.7 $\pm$ 0.2 \\
$T^{cir}$ & $P^{con}$ & 
\cellcolor[HTML]{B2CDFB}72.1 $\pm$ 0.6 & \cellcolor[HTML]{79A9F8}75.5 $\pm$ 0.3 & \cellcolor[HTML]{89B3F9}75.0 $\pm$ 1.0 & \cellcolor[HTML]{BED5FC}70.6 $\pm$ 0.5 & \cellcolor[HTML]{6A9FF7}85.4 $\pm$ 0.1 & \cellcolor[HTML]{4285F4}86.0 $\pm$ 0.1 & \cellcolor[HTML]{5F97F6}85.6 $\pm$ 0.2 & \cellcolor[HTML]{72A4F8}85.3 $\pm$ 0.3 \\
$T^{cir}$,$T^{sc2}$ & $P^{con}$ &  \cellcolor[HTML]{BCD4FC}70.9 $\pm$ 0.6 & \cellcolor[HTML]{659CF7}76.1 $\pm$ 0.7 & \cellcolor[HTML]{5491F6}76.6 $\pm$ 0.8 & \cellcolor[HTML]{AAC8FB}73.1 $\pm$ 0.1 & \cellcolor[HTML]{BCD4FC}83.9 $\pm$ 0.3 & \cellcolor[HTML]{B1CDFB}84.2 $\pm$ 0.1 & \cellcolor[HTML]{BFD6FC}83.8 $\pm$ 0.2 & \cellcolor[HTML]{E4EEFE}82.9 $\pm$ 0.1 \\
\hline
TripletWeight & PairWeight & lr 0.2 & lr 0.4 & lr 0.6 & lr0.8 & lr 0.5 & lr 1.0 & lr 2.0 & lr 4.0 \\
\hline
$T^{con}$ & $P^{lin}$                 & \cellcolor[HTML]{B3CEFB}77.2 $\pm$ 0.5 & \cellcolor[HTML]{97BCFA}77.8 $\pm$ 0.9 & \cellcolor[HTML]{B9D2FC}77.0 $\pm$ 0.6 & \cellcolor[HTML]{D5E4FD}76.3 $\pm$ 0.6 & \cellcolor[HTML]{BAD2FC}86.2 $\pm$ 0.1 & \cellcolor[HTML]{7CAAF8}87.2 $\pm$ 0.3 & \cellcolor[HTML]{6EA2F7}87.3 $\pm$ 0.2 & \cellcolor[HTML]{ACC9FB}86.5 $\pm$ 0.4 \\
$T^{cos}$ & $P^{lin}$                 & \cellcolor[HTML]{CADDFD}76.6 $\pm$ 0.8 & \cellcolor[HTML]{AAC8FB}77.4 $\pm$ 0.5 & \cellcolor[HTML]{A7C6FB}77.5 $\pm$ 0.5 & \cellcolor[HTML]{AECAFB}77.3 $\pm$ 0.1 & \cellcolor[HTML]{BFD6FC}86.0 $\pm$ 0.2 & \cellcolor[HTML]{6BA0F7}87.4 $\pm$ 0.2 & \cellcolor[HTML]{4285F4}87.9 $\pm$ 0.4 & \cellcolor[HTML]{6099F6}87.5 $\pm$ 0.3 \\
$T^{cos}$,$T^{sc1}$ & $P^{lin}$ & 
\cellcolor[HTML]{D9E7FD}76.2 $\pm$ 0.2 & \cellcolor[HTML]{4788F5}78.8 $\pm$ 0.6 & \cellcolor[HTML]{5994F6}78.6 $\pm$ 0.6 & \cellcolor[HTML]{84B0F9}78.0 $\pm$ 0.3 & \cellcolor[HTML]{CADDFD}85.8 $\pm$ 0.2 & \cellcolor[HTML]{99BDFA}86.8 $\pm$ 0.2 & \cellcolor[HTML]{6FA2F7}87.3 $\pm$ 0.3 & \cellcolor[HTML]{A3C4FB}86.7 $\pm$ 0.3 \\
$T^{cir}$ & $P^{lin}$                 & \cellcolor[HTML]{FFFFFF}75.2 $\pm$ 0.2 & \cellcolor[HTML]{B2CDFB}77.2 $\pm$ 0.2 & \cellcolor[HTML]{6BA0F7}78.3 $\pm$ 0.2 & \cellcolor[HTML]{BDD4FC}76.9 $\pm$ 0.4 & \cellcolor[HTML]{D4E3FD}85.5 $\pm$ 0.1 & \cellcolor[HTML]{93B9FA}86.9 $\pm$ 0.1 & \cellcolor[HTML]{5390F6}87.7 $\pm$ 0.2 & \cellcolor[HTML]{639AF7}87.5 $\pm$ 0.3 \\
$T^{cir}$,$T^{sc2}$ & $P^{lin}$ & 
\cellcolor[HTML]{F8FBFF}75.4 $\pm$ 0.6 & \cellcolor[HTML]{86B1F9}78.0 $\pm$ 0.4 & \cellcolor[HTML]{4285F4}78.8 $\pm$ 0.3 & \cellcolor[HTML]{5F98F6}78.5 $\pm$ 0.3 & \cellcolor[HTML]{FFFFFF}84.5 $\pm$ 0.1 & \cellcolor[HTML]{B7D1FC}86.2 $\pm$ 0.1 & \cellcolor[HTML]{9EC0FA}86.8 $\pm$ 0.2 & \cellcolor[HTML]{92B9FA}86.9 $\pm$ 0.2 \\
\hline
\end{tabular}
\end{adjustbox}
\caption{Comparing recall@1 performance of different triplet weights on CAR and In-shop dataset with various learning rates(lr).}
\label{table:grad_wt_raw}
\end{table*}

\section{Raw data table of gradient component}
Appendix Table~\ref{table:grad_gd_raw} shows more raw data with various learning rates for isolated effect of gradient direction. Appendix Table~\ref{table:grad_wp_raw} shows more raw data with various learning rates for isolated effect of pair weight. Appendix Table~\ref{table:grad_wt_raw} shows more raw data with various learning rates for isolated effect of triplet weight.

\section{Analysis of relative-similarity term in MS gradient}
There are two terms in MS loss dynamically changing the pair weight. The self-similarity term has been discussed in sigmoid pair weight $P^{sig}$ from the main paper in Section 3.3 As for the relative-similarity term, the major effect is to increase or decrease the maximum magnitude of the pair weight. 

Given a negative pair, when its relative-similarity term $m_{-}^{sig}>1$, this indicates the selected negative example is relatively closer to anchor compared to other negative examples.  Then, the negative weight increases because the relative term decreases the denominator in $P_{-}^{sig-ms}$. When its relative-similarity term $m_{-}^{sig}<1$, indicating the selected negative example is relatively far away from anchor comparing to other negative examples, the negative weight decreases because the relative term increases the denominator in $P_{-}^{sig-ms}$. The latter situation will not exist under the training with hard negative mining. 

Given a positive pair, when its relative-similarity term $m_{+}^{sig}>1$, this indicates the selected positive example is relatively close to anchor compared to other positive examples, the positive weight decreases because the relative term increases the denominator in $P_{+}^{sig-ms}$. When its relative-similarity term $m_{+}^{sig}<1$, indicating the selected positive example is relatively far away from anchor comparing to other positive examples, the positive weight increases because the relative term decreases the denominator in $P_{+}^{sig-ms}$. The latter situation will not exist under the training with easy positive mining.

\begin{figure}[t]
    \centering
    \includegraphics[width=.49\columnwidth]{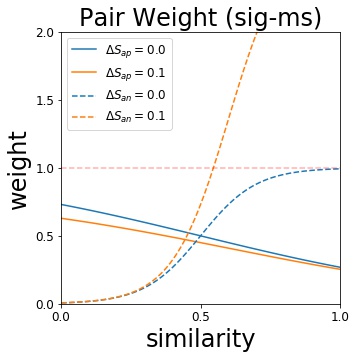}
    \includegraphics[width=.49\columnwidth]{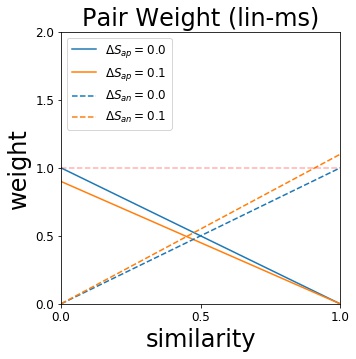}
    \caption{Comparison of sig-ms pair weight and lin-ms pair weight with relative similarity term $\Delta S_{ap}^{i} = \Delta S_{an}^{j}=0 \text{ and } 0.1$ when $\alpha=2$, $\beta=10$ and $\lambda=0.5$}
    \label{fig:exp_wp_sig2lin}
\end{figure}
In sum, the main effect caused by the relative-similarity term is to dynamically change the maximum penalty for positive and negative pairs as shown in Appendix Figure~\ref{fig:exp_wp_sig2lin}.

\section{Calculation instability of original MS weight}
The training result in Appendix Table~\ref{table:grad_wp_raw} reflects an apparent instability problem. This is caused by the exponential item $m_{-}^{sig}$ with a large $\beta$ which leads dramatically large value change that decreases the computation stability during the training.  

Appendix Figure~\ref{fig:exp_wp_sig2lin} shows the pair weight functions from Equation 13 and Equation 14 in the main paper with two different sets of relative similarity terms. Because the Easy Positive and Hard Negative(EPHN) mining~\cite{Xuan_2020_WACV} strategy is applied in training, each anchor has its positive and negative with the largest similarity. The relative similarity terms are always greater than zero, $\Delta S_{ap}^{i} = S_{ap}^s-{S_{ap}^r}^{i} \geq 0$ and $\Delta S_{an}^{j} = S_{an}^s-{S_{an}^r}^{j}\geq0$ (If all $\Delta S_{ap}^{i} = \Delta S_{an}^{j}=0$, the weight function degrade back to Equation 10 and 11 in the main paper). In addition, we set $\Delta S_{ap}^{i} = \Delta S_{an}^{j}=0.1$ in Appendix Figure~\ref{fig:exp_wp_sig2lin}.  

We see that sig-ms weight changes dramatically on the negative pair due to the exponential with the large $\beta$, generating an unstable gradient during the training. The problem occurs when the learning rate is large as shown in Appendix Table~\ref{table:grad_wp_raw}. With a small learning rate, especially in CAR tests, the sig-ms pair weight performs better than sig pair weight. But in In-shop tests, the unstable computation problem occurs and the performance of sig-ms pair weight is much worse than sig pair weight. 

In contrast, with the relative similarity term, the variation of our lin-ms pair weight is well controlled for both positive and negative pairs as shown in Appendix Figure~\ref{fig:exp_wp_sig2lin}. Also, in Appendix Table~\ref{table:grad_wp}, the performance of lin-ms pair weight performs better and is more stable with respect to learning-rate than sig-ms pair weight. 


\section{Implementation difference to the origin MS and DR-MS work}
Because both original MS~\cite{wang2019multi} and DR-MS~\cite{Mohan_2020_CVPR} work have not, to our knowledge, released complete training details for their SOTA results, including batch size and learning rate. We try our best to re-implemente  these two methods but also keep a fair comparison with other algorithms. 

The re-implementation of sigmoid pair weight and MS pair weight uses hyper-parameters $\alpha=2$, $\beta=10$, $\lambda=0.5$ and $\epsilon=0.1$. This setting is similar to the origin MS paper and other implementations such as DR-MS~\cite{Mohan_2020_CVPR} paper and Reality Check~\cite{musgrave2020metric} paper. We set $\lambda=0.5$ to make the weight symmetric when similarity varies from 0 to 1 and support fair comparison with linear pair weight which are also symmetric when similarity varies from 0 to 1. 

In Table 5 of the main paper, this re-implementation outperforms what is originally reported by the MS paper on CUB, CAR, and SOP with batch size 128. However, in MS paper, the best recall@1, $78.2\%$ of SOP uses batch size 1000. In the DR-MS paper, the best recall@1 of In-shop, $91.7\%$, use batch size 600. When DR-MS uses batch size 160 (a comparable batch size to our setting), recall@1 is $88.3\%$.

\section{The gradient projection length along $d_{\bot}$ for $e^{euc}$ and $e^{cos}$}
Let Euclidean gradient projection length along positive feature vector be $l_{\parallel pos}^{euc}$, along negative feature vector be $l_{\parallel neg}^{euc}$. 
\begin{equation}
\left\{
\begin{aligned}
&l_{\parallel pos}^{euc}=\mathbf{e_{p}^{euc}}\mathbf{f_p}=\frac{(\mathbf{f_p}-\mathbf{f_a})\mathbf{f_p}}{\|\mathbf{f_p}-\mathbf{f_a}\|}=\frac{1-S_{ap}}{\sqrt{2-2S_{ap}}}\\
&l_{\parallel neg}^{euc}=\mathbf{e_{n}^{euc}}\mathbf{f_n}=\frac{(\mathbf{f_a}-\mathbf{f_n})\mathbf{f_n}}{\|\mathbf{f_a}-\mathbf{f_n}\|}=\frac{S_{an}-1}{\sqrt{2-2S_{an}}}
\end{aligned} 
\right.
\end{equation}

The the orthogonal projection length $l_{\bot pos}^{euc}$ and $l_{\bot neg}^{euc}$ are:
\begin{equation}
\left\{
\begin{aligned}
&l_{\bot pos}^{euc}=\sqrt{1-{l_{\parallel pos}^{euc}}^2}=\sqrt{\frac{1+S_{ap}}{2}}\\
&l_{\bot neg}^{euc}=\sqrt{1-{l_{\parallel neg}^{euc}}^2}=\sqrt{\frac{1+S_{an}}{2}}
\end{aligned} 
\right.
\end{equation}

Let cosine gradient projection length along positive feature vector be $l_{\parallel pos}^{cos}$, along negative feature vector be $l_{\parallel neg}^{cos}$. 

\begin{equation}
\left\{
\begin{aligned}
&l_{\parallel pos}^{cos}=\mathbf{e_{p}^{cos}}\mathbf{f_p}=-\mathbf{f_a}\mathbf{f_p}=-S_{ap}\\
&l_{\parallel neg}^{cos}=\mathbf{e_{n}^{cos}}\mathbf{f_n}= \mathbf{f_a}\mathbf{f_m}=S_{ap}
\end{aligned} 
\right.
\end{equation}

The the orthogonal projection length $l_{\bot pos}^{cos}$ and $l_{\bot neg}^{cos}$ are:
\begin{equation}
\left\{
\begin{aligned}
&l_{\bot pos}^{cos}=\sqrt{1-{l_{\parallel pos}^{cos}}^2}=\sqrt{1-S_{ap}^2}\\
&l_{\bot neg}^{cos}=\sqrt{1-{l_{\parallel neg}^{cos}}^2}=\sqrt{1-S_{an}^2}
\end{aligned} 
\right.
\end{equation}